\documentclass{article}

\PassOptionsToPackage{numbers}{natbib}
\usepackage[final]{neurips_2024}
\usepackage[utf8]{inputenc} % allow utf-8 input
\usepackage[T1]{fontenc}    % use 8-bit T1 fonts
\usepackage{url}            % simple URL typesetting
\usepackage{booktabs}       % professional-quality tables
\usepackage{tabularx}       % for more flexible table widths
\usepackage{amsmath}
\usepackage{amsfonts}       % blackboard math symbols
\usepackage{nicefrac}       % compact symbols for 1/2, etc.
\usepackage{microtype}      % microtypography
\usepackage{xcolor}         % colors
\usepackage{graphicx}
\usepackage{wrapfig}
\usepackage{bbm}
\usepackage{multicol}
\usepackage[colorlinks=true,citecolor=cyan]{hyperref}
\definecolor{bo}{HTML}{BF5700}

\title{AMAGO-2: Breaking the Multi-Task Barrier in Meta-Reinforcement Learning with Transformers}

\author{%
Jake Grigsby \hspace{-2mm} \quad Justin Sasek$^{\dagger}$ \quad \hspace{-2mm}Samyak Parajuli$^{\dagger}$ \quad 
 \hspace{-2mm} Daniel Adebi$^{\dagger}$ \quad \hspace{-2mm} \textbf{Amy Zhang} \quad \hspace{-2mm}
 \textbf{Yuke Zhu} \\
The University of Texas at Austin \\
$\dagger$ Equal contribution \\
\texttt{\{grigsby,yukez\}@cs.utexas.edu}\\
}

\begin{document}

\maketitle

\begin{abstract}
Language models trained on diverse datasets unlock generalization by in-context learning. Reinforcement Learning (RL) policies can achieve a similar effect by meta-learning within the memory of a sequence model. However, meta-RL research primarily focuses on adapting to minor variations of a single task. It is difficult to scale towards more general behavior without confronting challenges in multi-task optimization, and few solutions are compatible with meta-RL's goal of learning from large training sets of unlabeled tasks. To address this challenge, we revisit the idea that multi-task RL is bottlenecked by imbalanced training losses created by uneven return scales across different tasks. We build upon recent advancements in Transformer-based (in-context) meta-RL and evaluate a simple yet scalable solution where both an agent's actor and critic objectives are converted to classification terms that decouple optimization from the current scale of returns. Large-scale comparisons in Meta-World ML45, Multi-Game Procgen, Multi-Task POPGym, Multi-Game Atari, and BabyAI find that this design unlocks significant progress in online multi-task adaptation and memory problems without explicit task labels.
\end{abstract}

\section{Introduction}
\label{sec:intro}

Billion-parameter generative models trained to imitate human language and behaviors from web datasets have unlocked unprecedented generality in machine learning \cite{touvron2023llama, team2023gemini}. While evaluations of these large language models (LLMs) are organized into discrete benchmarks \cite{hendrycks2020measuring} with specific themes like competitive mathematics \cite{Hendrycks2021MeasuringMP} or reading comprehension \cite{dua-etal-2019-drop}, their knowledge transfers to countless problems. LLMs' flexibility is largely due to an “in-context learning” effect that emerges when input sequences grow long enough to resemble a dataset of examples for a new task \cite{brown2020language, kirsch2022general}. However, predicting the most likely continuation of a sequence is misaligned with optimal control and limits us to tasks where high-quality demonstrations are available.

\begin{figure}[h!]
    \centering
    \includegraphics[width=.95\textwidth]{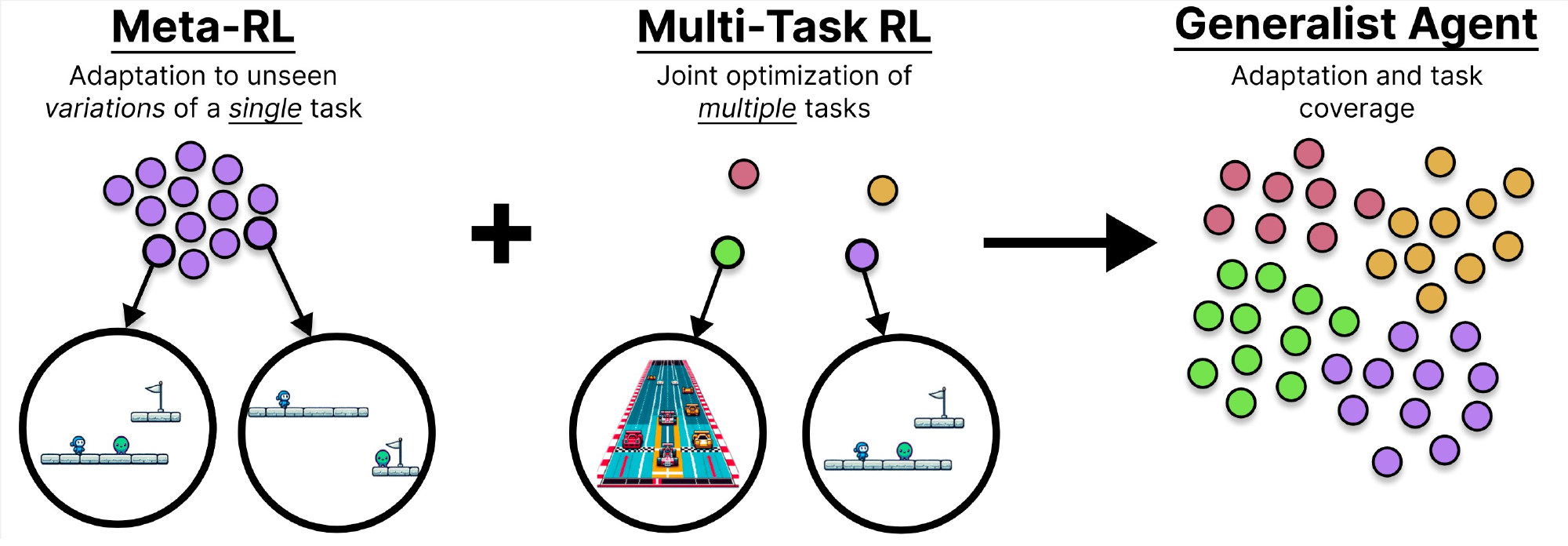}
    \caption{\textbf{Task Spaces in RL Generalization.} Meta-RL agentso adapt to dense variations of a core task. Multi-Task RL overcomes optimization challenges of learning from isolated tasks. Scalable ideas from both areas allow us to extend adaptive agents towards increasingly general behavior.\vspace{-4mm}}
    \label{fig:fig1}
\end{figure}

Online Reinforcement Learning (RL) lets models continuously self-improve and directly optimizes performance. The RL equivalent of in-context learning is a subset of meta-RL techniques that use sequence models for memory and adaptation \cite{beck2023survey, wang2016learning, duan2016rl}. Meta-RL can be viewed as a \textit{task inference} problem in which an agent explores its new surroundings to discover information that it can use to improve decision-making \cite{humplik2019meta, ghosh2021generalization, zintgraf2022fast}. In theory, meta-RL can adapt to a wide range of control problems. However, current applications are quite narrow ---  usually focusing on minor variations of what could be considered a single task. For example, imagine a platformer video game with procedurally generated levels. The game has a consistent theme and controls, but each level introduces a slightly different reward function, physics, and layout. A meta-learning policy trained on this game would learn to identify the relevant changes in a new level in order to maximize returns. By most formal definitions, any two levels from this game would be considered two different ``tasks,'' but it seems unlikely that anyone outside the field would refer to the resulting policy as a ``multi-task" agent. A truly general agent would also be able to play entirely different games. We will use the term ``task" to refer to this kind of multi-game generality while calling different levels of the same game ``task \textit{variants}” \cite{zhao2022effectiveness, taiga2023investigating, rusu2022probing}. Each task (game) creates a distribution of variants (levels) we can sample. The distinction between two ``tasks'' and two ``variants'' of the same task is arbitrary in theory\footnote{The distinction between tasks/variants is also called non-parametric/parametric variation \cite{yu2020meta, lee2023parameterizing}. Variants are often generated by randomizing a known set of parameters in a single (simulated) environment.}. However, it is so relevant in practice that having separate terms will be important for clarity.

Ideally, we would train RL agents on many different tasks and many variants of those tasks. Recent works have scaled to heterogeneous mixtures of popular RL benchmarks \cite{reed2022a, hansen2023td, gallouedec2024jack}, but rely on offline datasets collected by expert single-task agents. Scaling up online learning is challenging because multi-task distributions can create disconnected datasets; as we transition from a single task to small sets of several tasks, we reach a barrier where the gap between tasks becomes so wide that practical optimization challenges hinder learning more than any knowledge transfer between them is helping \cite{taiga2023investigating, shaham2023causes, xin2022current}. This conflict is particularly limiting in meta-RL because multi-task behavior should otherwise be a simple case of task inference. Meta-RL agents learn to infer the identity of task variants based on limited information --- which becomes more challenging as those variants become more similar. Adding an entirely new task with clearly distinct visuals or controls does little to the difficulty of this problem, and yet it can dramatically reduce performance because it introduces the (mostly unrelated) challenge of optimizing a second reward function. Fortunately, this challenge is not unique to meta-RL and is a focus of broader research in multi-task RL (MTRL) and supervised learning \cite{zhang2021survey}. MTRL methods can improve multi-objective training by creating task-specific network parameters or editing the gradients of each task to resolve conflicting updates (Section \ref{sec:background}). Unfortunately, these solutions rely on our ability to manually label tasks, scale with the size of our training set, and can abandon meta-RL's goal of adapting to new tasks at test-time. While MTRL might be practical on current benchmarks, it would be more scalable to preserve meta-RL's task inference and improve multi-task optimization without introducing new assumptions (Figure \ref{fig:fig1}).

This paper focuses on training sequence model policies that generalize across hybrid task distributions containing elements of meta-RL, MTRL, and long-term memory. We revisit the idea that MTRL is primarily bottlenecked by imbalanced training losses created by tasks with returns on significantly different scales \cite{hessel2019multi}. However, we approach this problem from a meta-RL perspective where \textit{task inference needs to be learned; we do not allow knowledge of task labels to influence training in any way}. We experiment with Transformer \cite{vaswani2017attention} policies trained by mixtures of actor and critic objectives that are either directly or indirectly dependent on the scale of returns in any given task. Comparisons on Meta-World ML45 \cite{yu2020meta}, Multi-Task POPGym \cite{morad2023popgym}, Multi-Game Procgen \cite{cobbe2020leveraging}, Multi-Game Atari \cite{bellemare2013arcade}, and Multi-Task BabyAI \cite{chevalier-boisvert2018babyai} evaluate the importance of scale-resistant updates. We find that converting value regression to classification \cite{hafner2023mastering} and policy improvement to filtered imitation learning \cite{wang2020critic} leads to significant improvements in these hybrid settings. Our results suggest that scale invariance should be a priority when designing multi-task agents with adaptive memory. Our empirical success with classification updates further supports recent observations that sequence-based RL can improve by borrowing technical details from supervised sequence modeling \cite{chebotar2023q, farebrother2024stop, springenberg2024offline} while retaining the online self-improvement process that makes RL useful \cite{chen2021decision, lee2022multi}.

\section{Background}
\label{sec:background}

\paragraph{Memory-Based Meta-RL.} The objective of a meta-RL agent is to adapt its policy to maximize returns over a horizon of one or more attempts in a new task. We focus on a subset of ``black-box'' approaches that reduce meta-learning to the problem of training sequence model policies with end-to-end RL \cite{beck2023survey}. We view adaptation as a form of partial observability \cite{ghosh2021generalization}, where instead of inferring the current state $s_t$ from a sequence of observations $(o_0, \dots, o_t)$, we try to infer the state \textit{and} the identity of the task from trajectories ($\tau$) of observations, actions ($a$), and rewards ($r$): $\tau_{0:t} = (o_0, a_0, r_0, o_1, \dots, a_{t-1}, r_{t-1}, o_t)$ \cite{zintgraf2022fast}. A task's identity is an unobservable variable describing aspects of its dynamics that vary across the training distribution \cite{hallak2015contextual}. Because improved estimates of the current task would increase returns, meta-reasoning can arise implicitly in the latent space of a sequence model that is trained by standard RL updates. 

This sequence-based meta-RL framework, which began with RL$^2$ \cite{wang2016learning, duan2016rl, botvinick2019reinforcement}, has two key advantages. First, training policies on trajectory sequences turns standard partial observability and zero-shot generalization \cite{kirk2022survey} into special cases. In practice, this creates flexible agents that blur formal boundaries between memory, generalization, and meta-learning \cite{benjamins2022contextualize}. Second, end-to-end RL enables self-improvement without demonstrations. Many meta-RL algorithms improve performance in applications where task identification could benefit from additional assumptions \cite{humplik2019meta, liu2021decoupling, rakelly2019efficient, beck2024splagger}. Another line of work recovers the flexibility of ``in-context'' sequence learning when a dataset of demonstrations makes self-improvement unnecessary \cite{laskin2022context, lee2024supervised, raparthy2023generalization, shi2024cross}. Most of the algorithmic variety in meta-RL comes from these two areas. In contrast, the RL$^2$ approach has primarily become an engineering problem. Important developments include increasing the adaptation horizon by enabling long-context sequence models in an RL setting \cite{mishra2017simple, melo2022transformers, lu2023structured}, and a shift away from on-policy updates in favor of off-policy actor-critics \cite{fakoor2019meta, yang2021recurrent, ni2022recurrent}. Combinations of these details allow Transformers with long-term recall to self-improve on large datasets of recycled data \cite{team2023human, ni2023transformers, grigsby2024amago}.

\paragraph{Multi-Task Optimization.} Multi-task learning addresses a trade-off where conflicting objectives decrease performance relative to single-task learning despite an increase in training data. One approach uses specialized optimizers that edit gradients to balance tasks' local optima \cite{shaham2023causes, du2018adapting, yu2020gradient, li2021robust, wang2020gradient, liu2021conflict}. However, it can be as effective (and less expensive) to rescale each task's contribution to the overall loss \cite{xin2022current, hessel2019multi, royer2024scalarization, kurin2022defense, sodhani2021multi}. These techniques require the ability to identify the current task, which is commonly available in multi-task settings: we can view MTRL as a meta-RL problem where the ground-truth identity of each environment has been one-hot labeled in advance. MTRL often depends on task labels in more subtle ways, including networks with task-specific output heads \cite{kumar2022offline} and datasets that balance sampling \cite{d2024sharing} or normalize rewards \cite{garage} across tasks. 

\paragraph{Problem Setting.} We focus on hybrid generalization problems that require elements of meta-learning, multi-task learning, and long-term memory. Meta-RL often evaluates performance over $k > 1$ attempts or episodes. A ``$k$-episode'' objective maximizes the \textit{total} return over all $k$ attempts and creates an exploration/exploitation trade-off at test-time \cite{zintgraf2022fast}. A ``$k$-\textit{shot}'' objective allows for an exploration window where only the final episode counts towards the agent's score \cite{liu2021decoupling, stadie2018some}. We measure generality by dividing a set of task variants ($\mathcal{V}$) into train and test sets. Meta-RL benchmarks are (informally) characterized by their relative density --- meaning $|\mathcal{V}|$ is large and composed of subtle changes to a common objective. Multi-task problems are created by training a single policy to maximize $N$ qualitatively distinct objectives, such as learning to play $N$ Atari games  \cite{bellemare2013arcade}. More general settings might be described as containing $N$ adaptation problems with their own set of variants $\mathcal{V}_0, \dots, \mathcal{V}_N$. These multi-task meta-learning problems are created by attempts to expand training sets by combining existing environments.

\paragraph{Scaling Meta-RL Beyond the Multi-Task Barrier.} Early meta-RL experiments focused on adapting to minor variations of multi-armed bandits \cite{wang2016learning}, gridworlds \cite{stadie2018some}, and locomotion benchmarks \cite{rakelly2019efficient, finn2017model, rothfuss2018promp}. Despite significant algorithmic progress, these domains are still representative of recent research \cite{beck2024splagger, laskin2022context, ni2022recurrent, rimon2024mamba}. Meta-RL requires generating many variations of a core problem while keeping those changes partially observed; it is difficult to design a cohesive benchmark that supports the dense level of variety it takes to induce meta-learning \cite{team2023human, wang2021alchemy}. We can create toy meta-RL problems, but there is a wide gap in complexity before we reach naturally occurring domains where long-term adaptation is necessary. For example, one application of meta-RL would be to learn to master unfamiliar video games over several minutes or hours. Building such an agent would likely involve scaling to a large training set of games. It is difficult to make gradual progress towards this goal because it begins as a memory problem, becomes a multi-task problem, and then only requires meta-learning at an extreme scale. Individual games often require long-term memory, but as we add games, there will inevitably be isolated groups with unique objectives that introduce challenges from MTRL. Eventually, we reach a level of training diversity where adaptation becomes critical and meta-learning emerges. The central issue considered by this paper is that multi-task optimization limits train-time performance at task counts far smaller than would be necessary for test-time generalization to new tasks. A promising direction is to take a method that is already capable of both memory and meta-learning and then find a way to break the barrier where progress is limited by multi-task optimization. However, if we want to adapt to many tasks, we should avoid solutions that scale with the number of training tasks or depend on our ability to identify them manually. 

This ``multi-task barrier'' impacts both long-term applications and current research benchmarks. For a more practical example, consider Meta-World \cite{yu2020meta}. Meta-World is a suite of $50$ robotic manipulation tasks. Each task creates a meta-RL problem by adjusting an unobservable goal so that agents must interpret incoming reward signals to adapt to the current objective. Meta-World ML45 extends generalization by merging meta-RL tasks into a multi-task training set $\cup_{i=0}^{N=45} \mathcal{V}_i$. Identifying the current task during training should be a special case of meta-RL because the changes between tasks are less ambiguous than the changes in goal location between variants of the same task. For example, an agent might distinguish a pick-and-place task from a door opening task after a single observation of the objects on the table. ML45 reserves five tasks to benchmark meta-RL's long-term goal of adapting to entirely new tasks at test-time. However, learning task adaptation from just $N=45$ training tasks is not realistic without additional assumptions \cite{yu2020meta, anand2022procedural}. ML45 is clearly bottlenecked by MTRL: a similar setup \textit{with the meta-RL component removed} is a challenging MTRL benchmark (MT50) by itself \cite{yu2020gradient, liu2021conflict}. MTRL methods use task labels to balance loss functions, separate network architectures, or edit task gradients. However, these techniques scale with $N$ and require a formal identification of each ``task'' that meta-RL otherwise should not need. Our goal is to overcome the same challenge while preserving the meta-RL perspective that task differences are arbitrary so that we can scale to unstructured domains where $N$ is large \cite{reed2022a} or infinite. Therefore, even though our experiments will have a clearly defined $N$, we will enforce a strict constraint where the agent can have zero knowledge of which task is active or how many tasks there are ($N$, $|\mathcal{V}_n|$).

\vspace{-3mm}
\section{Multi-Task Adaptation Without Task Labels}

\label{sec:method}

\begin{wrapfigure}{r}{.31\textwidth}
\vspace{-8mm}
\begin{center}
    \centering
    \includegraphics[width=.3\textwidth]{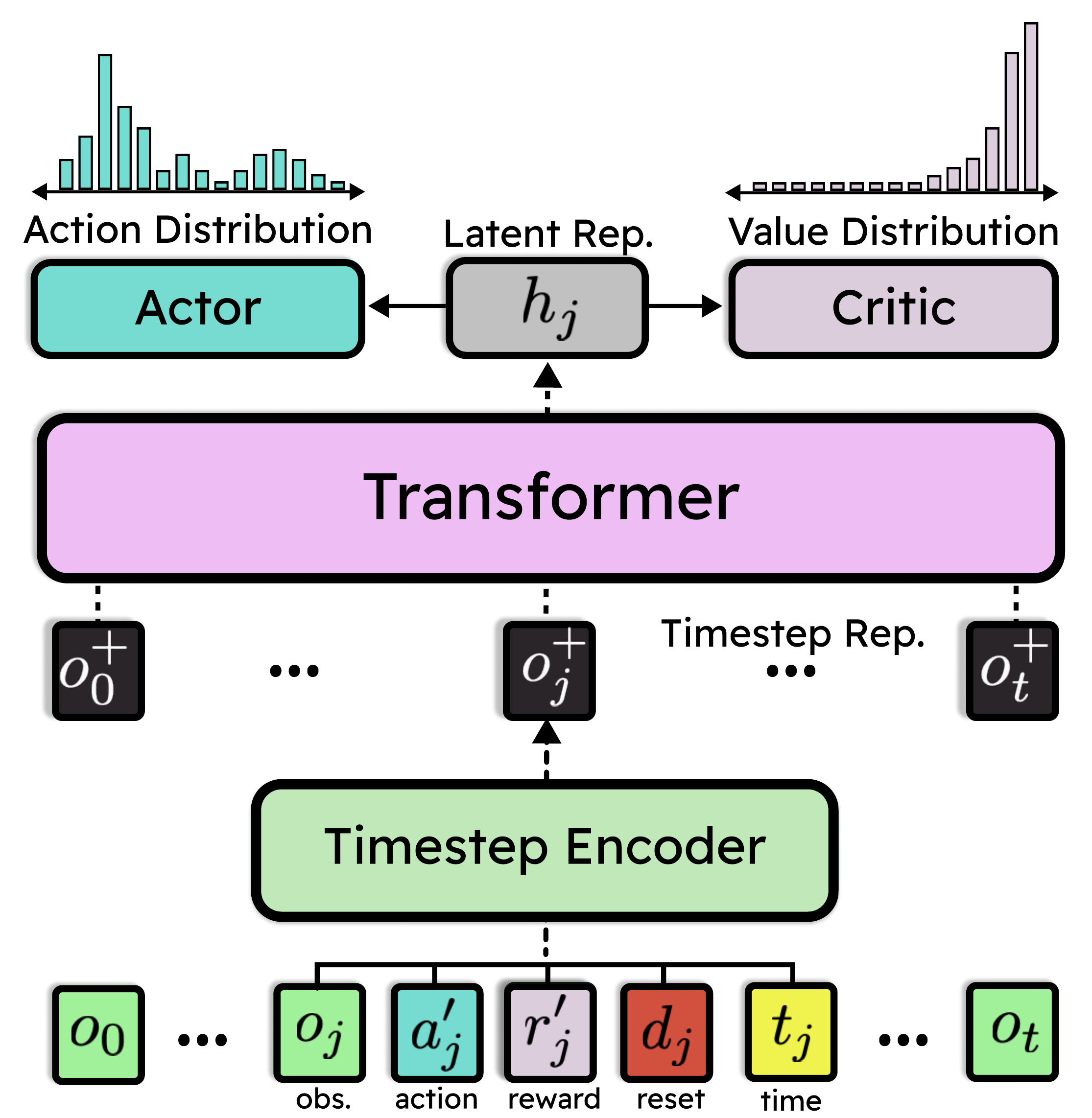}
    \vspace{-3mm}
    \caption{\textbf{Transformer-based Actor-Critic Architecture.}}
    \label{fig:arch}
\end{center}
\vspace{-6mm}
\end{wrapfigure}
We study multi-task adaptation without task labels by building on the flexible memory-based framework in which sequence models optimize standard RL objectives across trajectory inputs (Section \ref{sec:background}). Agents observe trajectory slices from a replay buffer with a context length of up to $l$ timesteps ($\tau_{t-l:t}$). The information revealed between timesteps ($o_i, a_{i-1}, r_{i -1}, d_{i -1})$ is merged into a single embedding $o^+_i$. The sequence $(o^+_{t-l}, \dots, o^+_t)$ is then passed through a sequence-to-sequence model to produce $(h_{t-l}, \dots, h_t)$ which serves as a shared representation for output heads representing the critic(s) $Q$ and stochastic policy $\pi$ (Figure \ref{fig:arch}). We focus on the combination of Transformers and off-policy RL updates, which will let us evaluate large-scale policies trained over millions of timesteps. Let $y_t = r_{t} + \gamma \bar{Q}(h_{t+1}, a' \sim \bar{\pi}(h_{t+1}))$ be the one-step temporal difference target, where $\bar{Q}$ and $\bar{\pi}$ refer to frozen target networks \cite{lillicrap2015continuous}. We can  minimize standard off-policy actor-critic loss terms \cite{lillicrap2015continuous, christodoulou2019soft} in parallel over the context length:

\begin{minipage}{0.45\textwidth}
  \begin{align}
    \label{eq:critic_dep}
    \mathcal{L}_{\text{Critic}}(t) &= \left(Q(h_t, a_t) - y_t\right)^2 
    \end{align}
\end{minipage}
\begin{minipage}{0.45\textwidth}
    \begin{align}
     \label{eq:actor_dep}
     \mathcal{L}_{\text{Actor}}(t) &= -Q\big(h_t, a \sim \pi(h_t)\big)
    \end{align}
\end{minipage}

Following details in AMAGO \cite{grigsby2024amago}, we compute Equations \eqref{eq:critic_dep} and \eqref{eq:actor_dep} with an ensemble of critics \cite{chen2021randomized} and over several different values of the discount factor $\gamma$ in parallel \cite{fedus2019hyperbolic}. By preventing the parameters of $Q$ from minimizing $\mathcal{L}_{\text{Actor}}$, we can update the full architecture with one step of the joint loss: $\sum_{i = t-l}^{t-1}\mathcal{L}_{\text{Actor}}(i) + \lambda \mathcal{L}_{\text{Critic}}(i)$, where $\lambda$ is a hyperparameter that balances the objectives for the shared sequence model \cite{ni2023transformers, grigsby2024amago}.

\begin{wrapfigure}{r}{.61\textwidth}
\vspace{-8mm}
\begin{center}
    \centering
    \includegraphics[width=.6\textwidth]{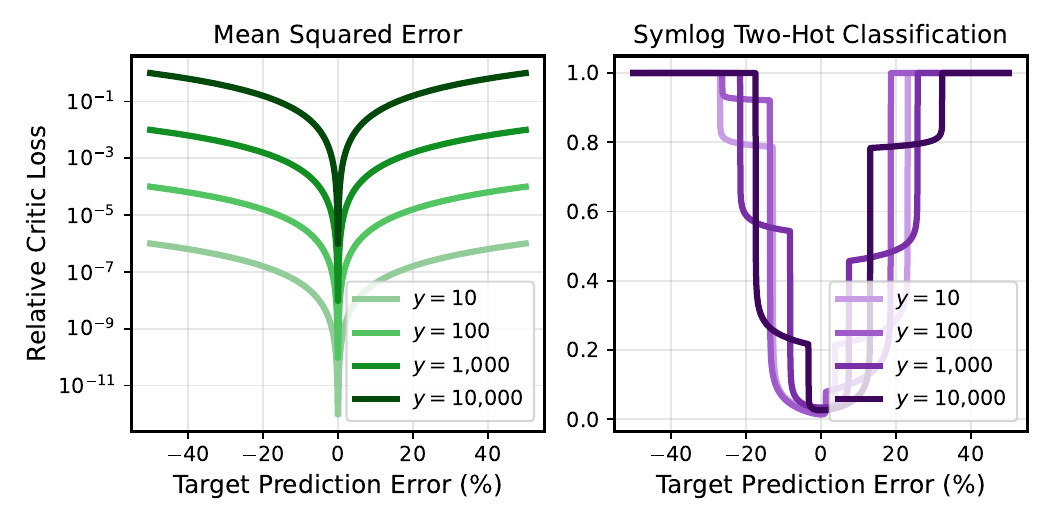}
    \vspace{-4mm}
    \caption{\textbf{Scale-Resistant Value Regression.} We plot the value of the standard critic loss (Eq. \ref{eq:critic_dep}) as a function of the \textit{relative} prediction error of the TD target ($y$) across four orders of magnitude (left). Y-axes are self-normalized according to the largest displayed value. Two-Hot classification (Eq. \ref{eq:actor_ind}) maps the same relative error to similar loss values across the different absolute return scales of each task (right).}
    \label{fig:relative_losses}
\end{center}
\vspace{-4mm}
\end{wrapfigure}

Multi-task optimization challenges are not unique to RL \cite{shaham2023causes, royer2024scalarization}, but RL typically adds a bottleneck that we can remove: both Equations \eqref{eq:critic_dep} and \eqref{eq:actor_dep} directly depend on the scale of returns $Q$ in each of our $N$ tasks. When computed as a simple expectation over tasks, the gradients of $\mathcal{L}_{\text{Critic}}$ and $\mathcal{L}_{\text{Actor}}$ can favor small \textit{relative} improvements in tasks with high \textit{absolute} returns (Figure \ref{fig:relative_losses} left). We can almost never assume tasks have similar return scales. For example, even when we design reward functions to have similar initial and optimal returns, differences in task difficulty lead to uneven learning progress and imbalanced $Q$-values. This issue was proposed and demonstrated by methods such as Multi-Task PopArt \cite{hessel2019multi}. PopArt's solution is to compute MTRL loss terms on a relative basis by creating a new critic output for each of the $N$ tasks. Each critic adaptively normalizes $y$ for its task and automatically rescales its output to help compensate for shifting targets. Simpler versions of this approach (e.g., without network rescaling) are universal in MTRL. However, we cannot use this technique without ground-truth knowledge of task labels. Although we cannot adaptively balance each task's loss, we can reformulate Eqs. \eqref{eq:critic_dep} and \eqref{eq:actor_dep} to be less sensitive to the scale of $Q$.

\paragraph{Scale-Resistant Critics.} Distributional RL \cite{bellemare2023distributional} is motivated by representation learning and risk-aware policies, but methods like C51 \cite{bellemare2017distributional} have a useful side-effect of turning \eqref{eq:critic_dep} into a classification problem that scales with the number of output bins instead of $Q$. For this reason, C51-style critic updates have become a key implementation detail in large-scale (offline) MTRL agents \cite{kumar2022offline, springenberg2024offline}. We can keep the classification side-effect without modeling return distributions by converting a scalar $Q$ to discrete label(s) $\in \{0, \dots, B\}$. Two-hot labels create a one-to-one mapping between every scalar in a fixed range and a discrete distribution where all non-zero probability is placed in two adjacent bins \cite{schrittwieser2020mastering, hessel2021muesli}. If $Q_B$ denotes a modified critic network that outputs probabilities over $B$ bins, the TD-error can be minimized by multi-label classification:

\vspace{-5mm}
\begin{align}
    \label{eq:critic_ind}
    \mathcal{L}_{\text{Critic-Ind}}(t) = -\text{twohot}_B(y_t)^\mathsf{T} \log Q_B(h_t, a_t)
\end{align}
\vspace{-5mm}

We can reduce tuning of label spacing and limits with an invertible transform that maps a wide range of returns to a smaller number of labels \cite{kapturowski2018recurrent}; we use \texttt{symlog} as demonstrated by DreamerV3 \cite{hafner2023mastering} and TD-MPC2 \cite{hansen2023td}. We also experimented with using a global (task-agnostic) $Q$-normalization for this purpose but found that a fixed mapping is important because stabilizing learning against shifting labels required extensive hyperparameter tuning. Figure \ref{fig:relative_losses} visualizes why we would expect classification to be an improvement in a multi-task setting: $\mathcal{L}_{\text{Critic-Ind}}$ maps the same relative error in value prediction to a similar loss value over a wide range return scales. In contrast, $\mathcal{L}_{\text{Critic}}$ can bias optimization towards tasks that happen to have large regression targets. This effect can be dependent on proper tuning of the bin count $B$, upper/lower return bounds, and use of the \texttt{symlog} transform (Appendix \ref{app:implementation}).

A concurrent work \cite{farebrother2024stop} evaluates the two-hot critic loss' ability to unlock scaling laws in RL when used as a direct replacement for $\mathcal{L}_{\text{Critic}}$ \eqref{eq:critic_dep} (or its algorithm-specific equivalent). They find the classification loss improves performance in a range of domains and attribute this to advantages in representation learning and robustness to noisy value targets ($y$). Our experiments focus on the PopArt hypothesis that scale-invariance improves optimization in an online multi-task setting. We will be exploring the impact of classification losses in an actor-critic agent with adaptive memory, where we also need to consider the scale of the policy's objective \eqref{eq:actor_dep}.

\paragraph{Scale-Resistant Actors.} Many policy updates resemble a weighted regression objective where the importance of each action label is scaled by a function $f(\tau, a)$ that increases with its value or advantage, such as $f(\tau, a) \propto \exp(A(\tau, a))$\footnote{Where $A(\tau, a) = Q(\tau, a) - Q(\tau, a' \sim \pi(\tau))$ is the improvement of $a$ over the value of the current policy as estimated by actor and critic networks that take trajectory sequences as input.}. Methods differ in where the candidate actions $a$ are sampled from, how values are estimated, and how $f(\tau, a)$ clips or rescales the update \cite{hessel2021muesli, schulman2017proximal, abdolmaleki2018maximum, peng2019advantage, wang2018exponentially}. In the off-policy setting where actions are sampled from a dataset of previous experience, slight differences in advantage estimates and weight functions $f$ create a dense family of similar advantage-weighted regression (AWR) methods \cite{peng2019advantage, wang2018exponentially, chen2020bail, siegel2020keep, grigsby2021closer}. AWR is intuitive and simple to implement but directly depends on the scale of returns unless we intentionally avoid this with a binary filter $f(\tau, a) = \mathbbm{1}\{A(\tau, a) > 0\}$ \cite{wang2020critic, nair2018overcoming}. We follow details from CRR-Binary-Mean \cite{wang2020critic} for a candidate actor loss that is independent of $Q$:

\vspace{-4mm}
\begin{align}
    \label{eq:actor_ind}
    \mathcal{L}_{\text{Actor-Ind}}(t) = -\mathbbm{1}\{Q(h_t, a_t) - \mathop{\mathbb{E}}_{a' \sim \pi(h_t)}[Q(h_t, a')] > 0\}\log\pi(a_t \mid h_t)
\end{align}
\vspace{-4mm}

Intuitively, CRR performs imitation learning (IL) on action labels that we estimate will improve the current policy. Advantage estimates within $\mathcal{L}_{\text{Actor-Ind}}$ rely entirely on one-step dynamic programming \eqref{eq:critic_dep} so that they never become outdated \cite{wang2020critic, nair2020awac}. Because they do not allow the policy to extrapolate to out-of-distribution actions, weighted IL updates like CRR are best suited to \textit{offline} RL \cite{levine2020offline}. In offline RL, the ability to stabilize optimization on static datasets is well worth losing the optimistic exploration of $\mathcal{L}_{\text{Actor}}$ \eqref{eq:actor_dep}. Choosing to use $\mathcal{L}_{\text{Actor-Ind}}$ online implies that the ability to address issues in multi-task optimization is a similarly worthwhile trade-off. There are other RL engineering factors to consider in the specific case of training long-context Transformers. For example, sampling batches of trajectory sequences can create replay ratios high enough to effectively be offline RL \cite{fedus2020revisiting}. However, the simplest justification for training Transformers on $\mathcal{L}_{\text{Actor-Ind}}$ may come from its similarities to widely used methods for supervised learning in RL.

\paragraph{Transformers and RL via Supervised Learning.} Decision Transformer (DT) is a popular approach to training sequence policies that reformulates RL as IL conditioned on a target return \cite{chen2021decision, janner2021reinforcement, lee2022multi}. DT's advantage is that it reduces to IL on expert datasets, making it a safe improvement over a technique that is already effective for many problems. In a field where engineering is critical, and results can be hard to reproduce, there is valuable simplicity in inheriting technical details from supervised sequence modeling. This simplicity has likely contributed to the method's success despite several disadvantages. For example, DT policies are conditioned on the optimal return in order to imitate the best continuation of the current trajectory, but $\mathcal{L}_{\text{Actor-Ind}}$ does this without needing to define the optimal return in advance. DT determines the value of an action based on the fixed return of its trajectory, while $\mathcal{L}_{\text{Actor-Ind}}$ continuously updates its estimate with the value of the current network. By extension, DT cannot handle stochastic reward functions \cite{paster2022you}. Like DT, $\mathcal{L}_{\text{Actor-Ind}}$ becomes IL on expert datasets, and Eqs.~\eqref{eq:critic_ind} and \eqref{eq:actor_ind} resemble supervised learning with two classification heads.

\section{Experiments}

$\mathcal{L}_{\text{Actor}}$, $\mathcal{L}_{\text{Critic}}$, $\mathcal{L}_{\text{Actor-Ind}}$, and $\mathcal{L}_{\text{Critic-Ind}}$ create four interchangeable combinations of learning updates where one or both of the actor and critic loss can be dependent or \textit{indirectly} dependent on the scale of $Q$. We will compare all four update variants, although we are mainly interested in evaluating the two extremes: ``Dep. Actor, Dep. Critic'' ($\mathcal{L}_{\text{Actor}}$, $\mathcal{L}_{\text{Critic}}$) and ``Ind. Actor, Ind. Critic'' ($\mathcal{L}_{\text{Actor-Ind}}$, $\mathcal{L}_{\text{Critic-Ind}}$). Our experimental setup allows for a direct ablation where all other details can be held fixed. Appendix \ref{app:implementation} includes additional implementation details. Our experiments focus on three main questions: \textbf{1)} Do scale-resistant actor-critic objectives offer empirical benefits in challenging generalization problems? \textbf{2)} If so, can these gains be attributed to multi-task return scaling? and \textbf{3)} What new applications of memory-based policies might be unlocked by improvements in multi-task optimization without task labels?

\begin{figure}[h!]
    \centering
    \includegraphics[width=\textwidth]{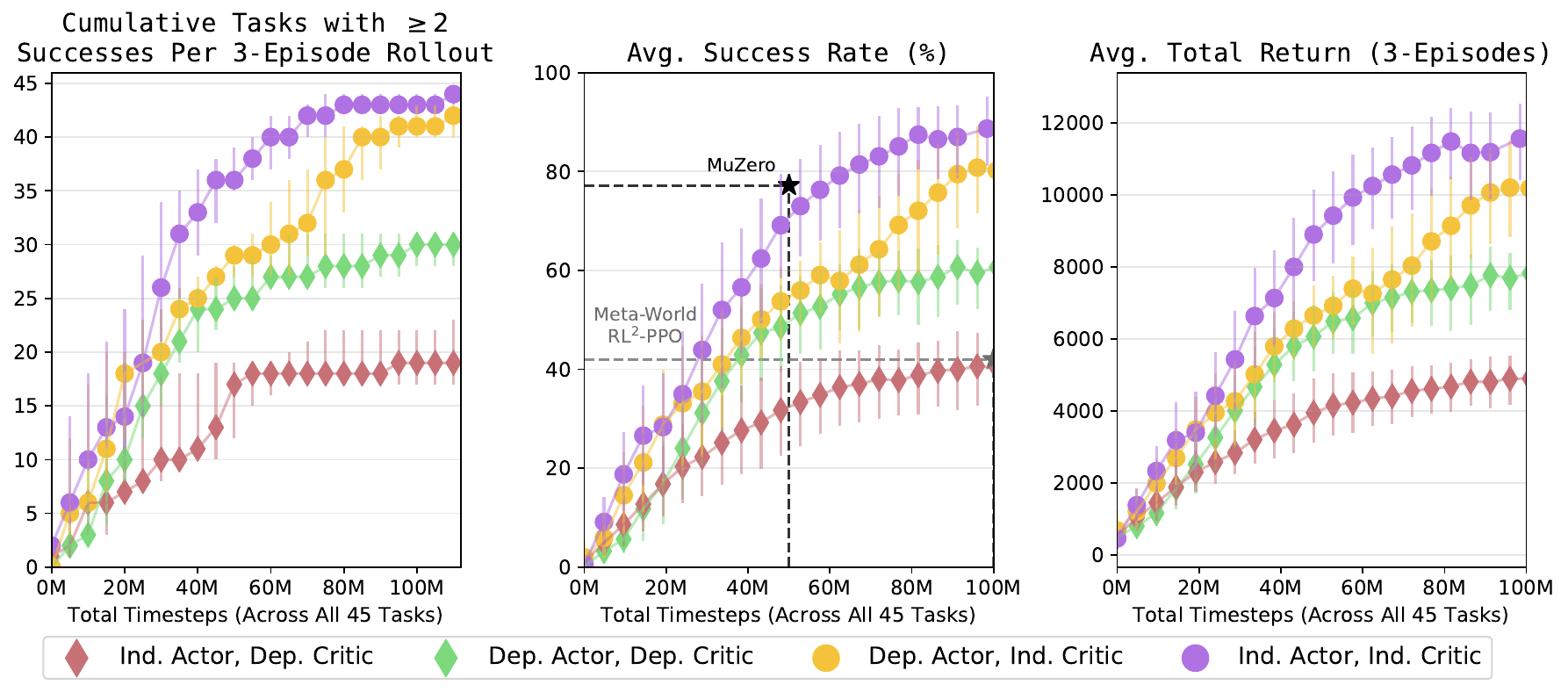}
    \caption{\textbf{Meta-World ML45 Train Task Results.} \textbf{(Left)} Coverage of the $45$ manipulation skills measured by an adaptation horizon success rate $\geq 2/3$. \textbf{(Center)} Average success rate over tasks, variants, and $3$-episode rollouts. Reference scores for MuZero and RL$^2$-PPO are gathered from results in \cite{yu2020meta, anand2022procedural}. \textbf{(Right)} Total return over a $3$-episode adaptation horizon, averaged across tasks and variants. All error bars indicate the maximum and minimum metric across three random trials.}
    \label{fig:metaworld_results}
    \vspace{-4mm}
\end{figure}

\paragraph{Meta-World ML45.} ML45 is a challenging suite of $45$ robotic manipulation meta-RL tasks. We argue that much of ML45's difficulty can be attributed to variance in return scales: ML45's training tasks are designed to have similar optimal returns, but they have different levels of difficulty, which causes the scale of $Q$-values to diverge early in training. The PopArt method of normalizing $Q$ per task may help balance optimization. Methods built on the original Meta-World results normalize the returns of each task separately \cite{garage}, which is a simpler approach to the same idea. We force all task identification to be learned within the context of our memory policy so that one-hot labels and the total task size ($45$) do not influence training. The training environment samples a new task between meta-rollouts and the task identity is not included in the observation. Every task shares a single critic head, and all experience is added to a shared replay buffer sampled uniformly at random. 

Figure \ref{fig:metaworld_results} compares the four combinations of learning updates described in Section \ref{sec:method}. Actor and critic loss terms that do not scale directly with the $Q$-values of each task improve skill coverage, overall success rate, and cumulative return across $3$-episode adaptation horizons. Replacing $Q$ regression with two-hot classification appears to be the most important change, although substituting the weighted IL policy update improves performance and sample efficiency. Full learning curves for all $45$ tasks are plotted in Appendix \ref{app:results} and highlight the uneven learning progress (and therefore $Q$-value scale) of tasks throughout training. With a sample budget of $100$M timesteps, our off-policy Transformers trained by max likelihood losses $(\mathcal{L}_{\text{Actor-Ind}}, \mathcal{L}_{\text{Critic-Ind}})$ more than double the success rate of Meta-World's original RL$^2$ result over the same sample limit while maintaining a $1.4\times$ improvement on its performance at $400$M timesteps. Our simple one-step $Q$-learning matches more complex value updates like MuZero \cite{schrittwieser2020mastering} at its reported $50$M timestep budget \cite{anand2022procedural}. We use a general Transformer architecture that would be equally applicable to any sequence modeling problem, but our results are also comparable to recent work such as HTrMRL \cite{shala2024hierarchical} (success rate $\approx 85
\%$) that use architectures specialized for multi-episodic RL. 

\begin{wrapfigure}{r}{.49\textwidth}
\vspace{-15mm}
\begin{center}
    \centering
    \includegraphics[width=.48\textwidth]{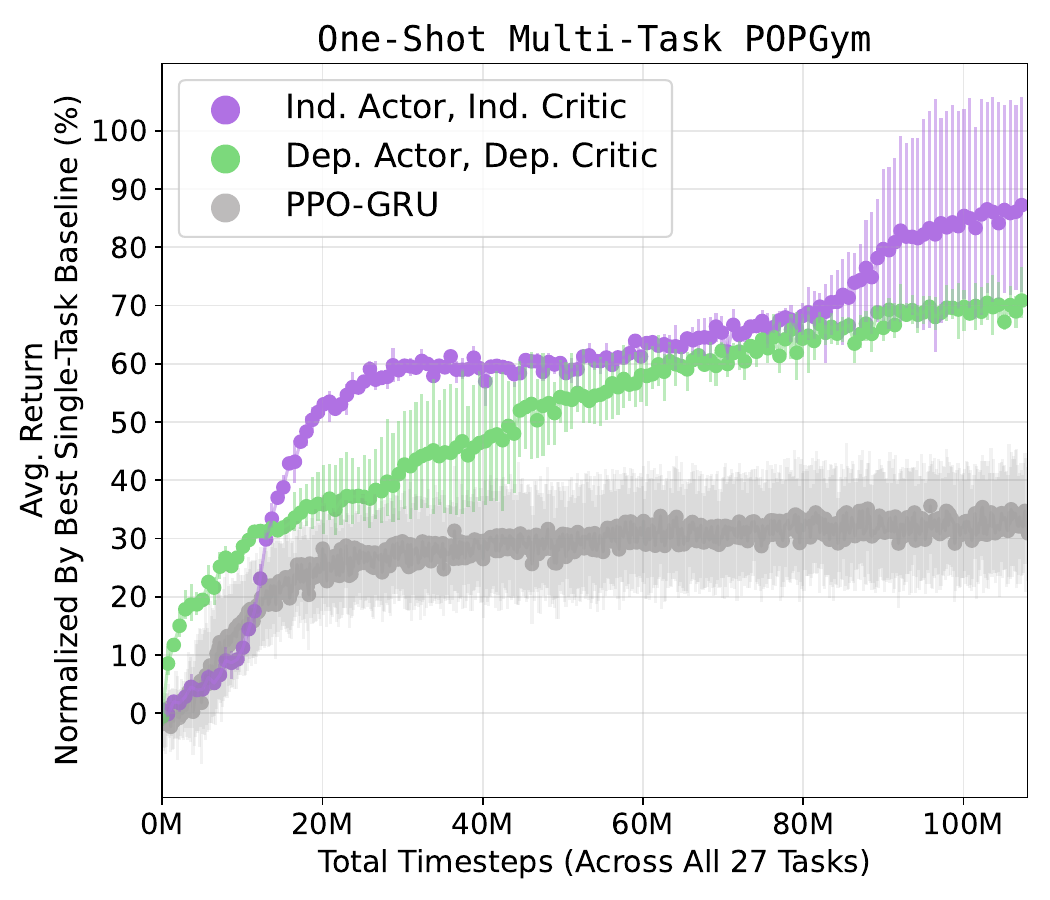}
    \vspace{-2mm}
    \caption{\textbf{Multi-Task POPGym Results.} Returns are normalized by single-task experts trained for $15$M timesteps. Error bars indicate the maximum and minimum returns across three trials.}
    \label{fig:popgym_results}
\end{center}
\vspace{-3mm}
\end{wrapfigure}
\paragraph{Multi-Task POPGym.} Meta-World is designed as a meta-learning benchmark, but its adaptation does not require long-term memory. POPGym \cite{morad2023popgym} is a suite of zero-shot POMDP problems. Many POPGym tasks are explicit tests of long-term recall, but others involve state and variant estimation from previous actions and rewards. We use a simple observation and action space padding approach to create a multi-task version of POPGym: we combine all $27$ tasks where both the action space and observation space have dimension $< 30$. It is not clear whether zero-shot performance in this setting can be compared against single-task POPGym results, as a strong agent may need to lose valuable time identifying which of the $27$ memory games it is currently playing. Therefore, we evaluate in a one-shot setting where the first trial is a free exploration window used to determine which task is active so that the second trial creates a fair comparison to single-task results \cite{stadie2018some}. Importantly, we avoid ``spoiling'' any memory challenges for the second trial by randomly resetting to a new variant of the same task.

We compare Transformers trained by the ``indirect'' and $Q$-dependent update against the on-policy PPO agent from the POPGym codebase \cite{morad2023popgym} using the best sequence model from the original benchmark (a GRU RNN \cite{cho2014learning}). Figure \ref{fig:popgym_results} aggregates the results across all $27$ tasks with returns scaled between a random policy and the best single-task result (of any sequence model) that appears in \citet{morad2024reinforcement}. Single-task reference scores are given a budget of $15$M timesteps, while our multi-task agents see a total of $100$M across all $27$ tasks and train on reward signals for approximately half of those timesteps. Both Transformer learning updates outperform PPO-GRU but are relatively comparable to each other. This may not be surprising because POPGym tasks have similar learning curves and returns bounded in $[-1, 1]$. However, many of these tasks have sharp short-term credit assignment where the advantage surface of an accurate value function would be steep. We expect the standard $\mathcal{L}_{\text{Actor}}$ update to be a significant improvement in online exploration, so it is encouraging that the offline-style $\mathcal{L}_{\text{Actor-Ind}}$ can exceed its performance after $15$M total timesteps. Learning curves for each task can be found in Appendix \ref{app:results}.

\begin{figure}[h!]
    \vspace{-2mm}
    \centering
    \includegraphics[width=.9\textwidth]{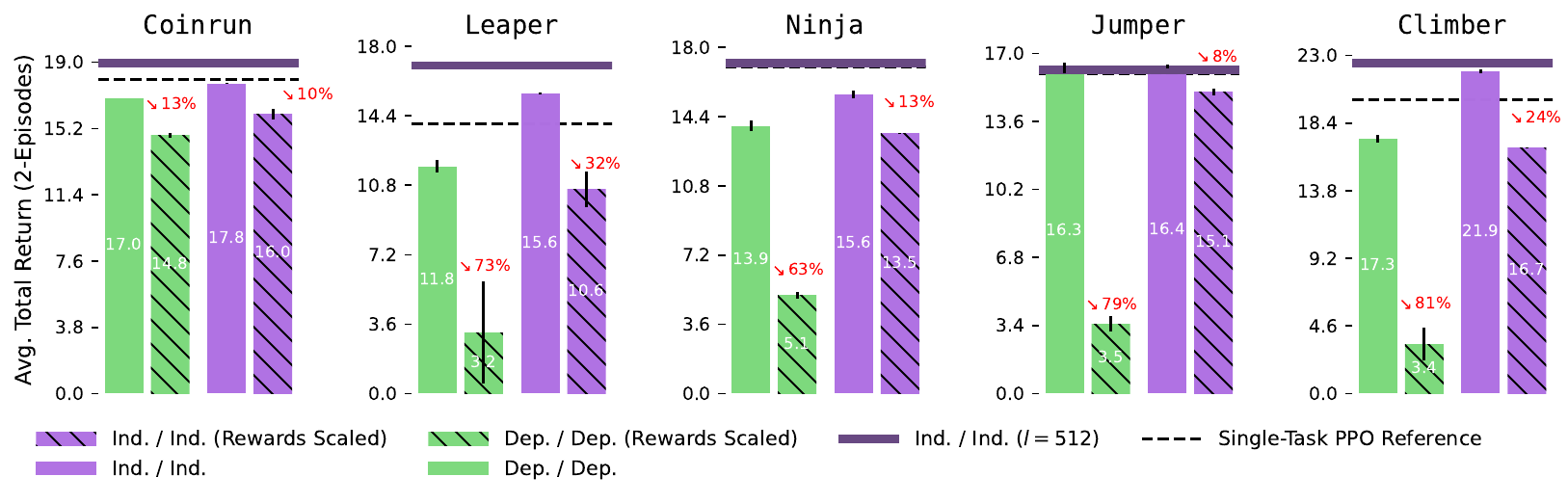}
    \vspace{-3mm}
    \caption{\textbf{Return Scales in Multi-Game Procgen.} We evaluate over $2$ episodes in unseen levels after optimizing default or rescaled reward functions. Scores are averaged over two model checkpoints and converted to the default scale. Error bars indicate the difference between two $275$M timestep trials.}
    \vspace{-2mm}
    \label{fig:popgym_easy_mode}
\end{figure}

\paragraph{Multi-Game Procgen.} Procgen \cite{cobbe2020leveraging} creates $16$ video games with an infinite variety of procedurally generated levels. If memory-based actor-critic learning is bottlenecked by the scale of returns across multiple tasks, we should be able to decrease performance by simply rescaling rewards. We begin by training multi-task agents across $2$k levels of five games in easy mode. Solid bars in Figure \ref{fig:popgym_easy_mode} show the performance on unseen test levels, where the scale-resistant update offers a slight improvement.

\begin{figure}[h!]
    \vspace{-3mm}
    \centering
    \includegraphics[width=\textwidth]{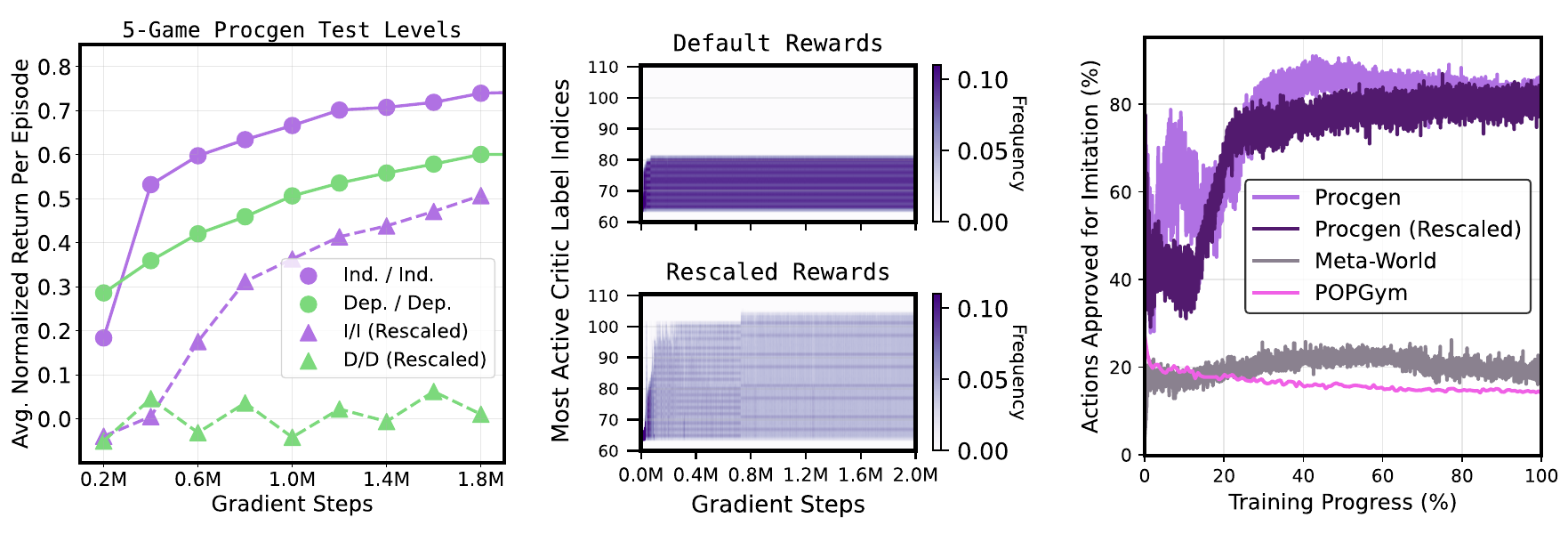}
    \caption{\textbf{Learning from Rescaled Returns.} \textbf{(Left)} Procgen test level returns normalized across games according to the standard benchmark scale \cite{cobbe2020leveraging}. \textbf{(Center)} Frequencies of the critic classification label index with the highest probability throughout training. We trim the y-axis to the portion of \texttt{symlog} \cite{hafner2023mastering} space that is relevant to Procgen. \textbf{(Right)} The value of $\mathcal{L}_{\text{Actor-Ind}}$ binary advantage weights create an estimate of the percentage of the replay buffer currently being imitated.}
    \label{fig:procgen_analysis}
\end{figure}

Our multi-task Procgen environment randomly selects a new game between resets, which does not account for how episode lengths depend on the agent's skill (Figure \ref{fig:procgen_frame_counts}). We retrain new policies from scratch but scale the rewards of the easiest task (Coinrun) by $\times 100$ while dividing the rewards of the task the agent spends the most timesteps interacting with (Climber) by $10$. We would expect the scale of $\mathcal{L}_{\text{Actor}}$ and $\mathcal{L}_{\text{Critic}}$ to be dominated by Coinrun so that optimization variance overshadows the objectives of the other environments even after learning in Coinrun converges. Hatched bars in Figure \ref{fig:popgym_easy_mode} show this is indeed the case, with methods recovering most of their original performance in Coinrun while failing to improve in the other four games. The ``Ind. / Ind.'' update is far less impacted by the new return scales even though they decrease efficiency in terms of gradient steps (Figure \ref{fig:procgen_analysis} Left). Fig. \ref{fig:procgen_analysis} (Center) reveals a simple explanation where the broader range of returns approximately doubles the number of frequently used labels our critics need to learn to classify. Multi-task Transformer policies are commonly trained by imitation learning, and Fig. \ref{fig:procgen_analysis} (Right) highlights why $\mathcal{L}_{\text{Actor-Ind}}$ is so effective for training multi-task Transformer policies with online RL: the policy update reduces to IL on a dynamic percentage of the replay buffer, but the ability to ignore a fraction of the dataset automatically allows for self-improvement in a way that standard IL does not.

\begin{figure}[h!]
    \vspace{-1mm}
    \centering
    \includegraphics[width=.9\textwidth]{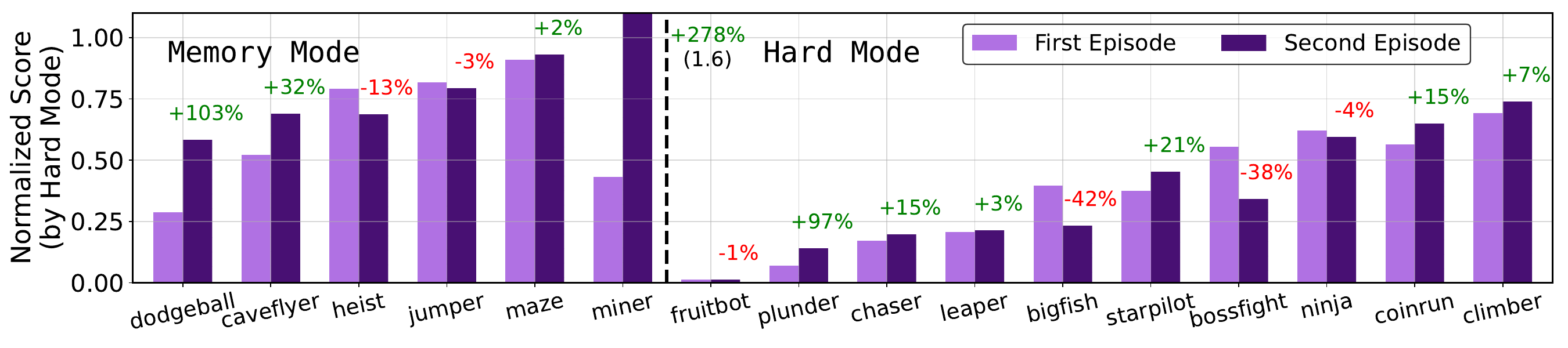}
    \vspace{-4mm}
    \caption{\textbf{Multi-Task Procgen in Memory-Hard Mode.} We measure policy performance across the two episodes of its adaptation window. Results are averaged over $30$M  frames in unseen test levels. }
    \label{fig:procgen_memory}
    \vspace{-1mm}
\end{figure}

Inspired by the success of the ``Ind. / Ind.'' learning update in the easy distribution of $5$ games, we train a larger $24$M parameter Transformer with a context length of $l=768$ frames on all $16$ Procgen games simultaneously. We default to the standard ``hard'' level distribution but increase the difficulty to the (rarely attempted) ``memory'' mode in the $6$ games it is available. Results after $2.7$B frames are highlighted in Figure \ref{fig:procgen_memory}. The agent does improve its performance on the second attempt of unseen levels --- particularly in memory games where partial observability makes adaptation most useful.

\begin{figure}[h!]
    \vspace{-2mm}
    \centering
    \includegraphics[width=.98\linewidth]{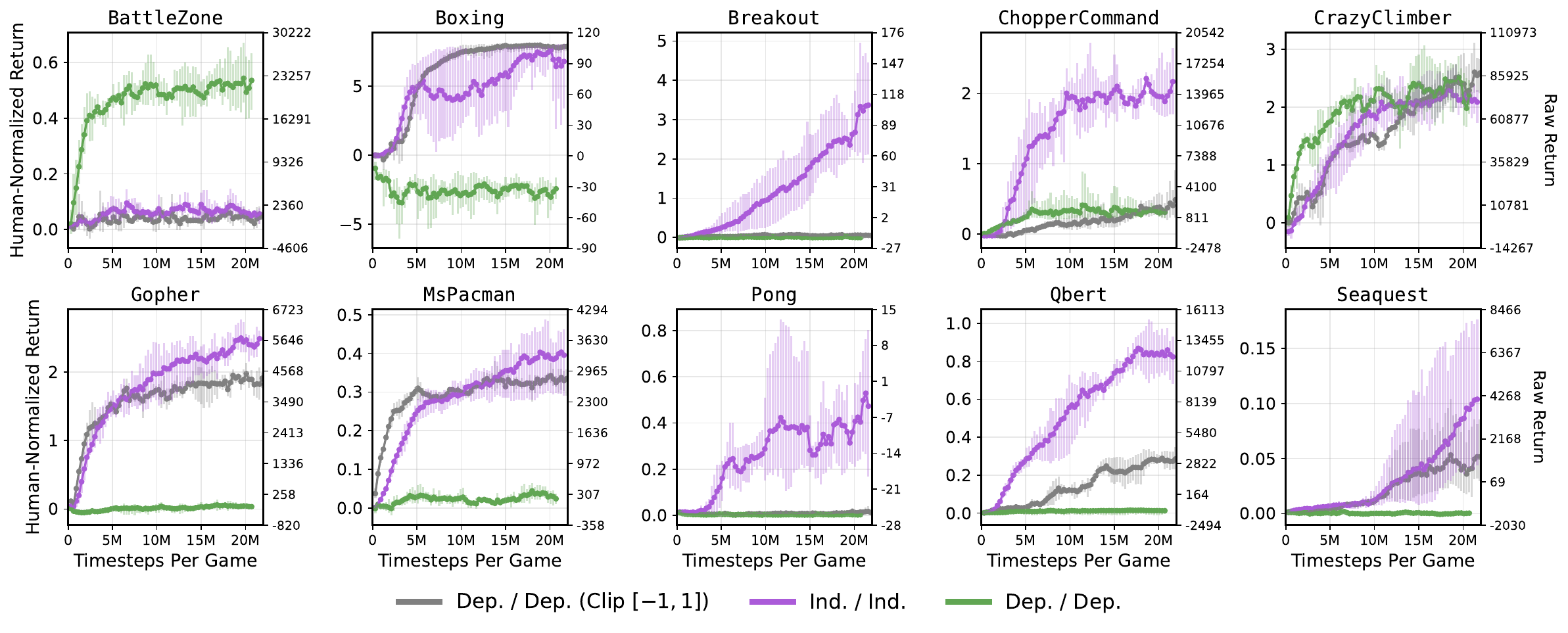}
    \vspace{-2mm}
    \caption{\textbf{Multi-Game Atari Without Reward Clipping.} We train a single policy on $10$ games simultaneously. Results are plotted relative to human performance (left axis) and the raw scale of returns (right axis). Error bars denote the maximum and minimum over four random trials.}
    \label{fig:atari_result}
    \vspace{-3mm}
\end{figure}

\paragraph{Multi-Game Atari.} If RL generalization problems lie on axes between meta-RL and MTRL (Section \ref{sec:background}), the ALE \cite{bellemare2013arcade} would represent the multi-task extreme. Super-human performance on Atari does not require adaptation or memory but leads to per-game returns separated by orders of magnitude \cite{hessel2018rainbow}, which motivates the infamous trick of clipping rewards in $[-1, 1]$ \cite{mnih2015human}; clipping stabilizes learning but misaligns agents' training objective from the raw game score used for evaluation. PopArt \cite{hessel2019multi} showed that multi-task training on unclipped rewards is possible when we split critic networks into game-specific output heads with independently normalized targets. Figure \ref{fig:atari_result} revisits this experiment on a smaller scale. However, rather than normalizing a multi-task architecture, we train a single critic on $10$ games with unclipped rewards and let our shared Transformer recover state and task labels implicitly from context sequences (of RGB images). In other words, we are treating the widely studied MTRL Atari setting as a special case of meta-learning in order to isolate the challenge of optimizing returns on different scales. Results are shown in Figure \ref{fig:atari_result}. The scale-resistant update leads to dramatic improvements in $8/10$ games; the default $Q$-dependent update performs best in the two games where returns at random initialization are significant outliers. The standard approach of clipping rewards in $[-1, 1]$ (grey) helps mask the multi-task optimization challenge created by scale-dependent learning updates.

\paragraph{BabyAI.} BabyAI is a suite of partially observed gridworld tasks with simple language instructions \cite{chevalier-boisvert2018babyai}. We create a hybrid multi-task meta-RL problem with the potential for task generalization by randomly generating a train/test split over $68$ of BabyAI's task configurations as provided by the Minigrid simulator \cite{MinigridMiniworld23}. Our agents observe a text description of the their task and have two attempts to adapt to a procedurally generated level layout. Learning curves for all $68$ tasks are provided in Appendix \ref{app:results}. Figure \ref{fig:babyai_main} summarizes our results in the BabyAI domain and reveals another case where scale-independent updates outperform scalar regression. Figure \ref{fig:babyai_icrl} evaluates BabyAI as a multi-episodic adaptation problem by comparing the returns of the first and second attempt in a new level layout. Despite BabyAI's partial observabilty, only a small set of tasks benefit from in-context adaptation over a context length $l=512$. Transformers are clearly capable of learning effective recall over sequences of this length via RL \cite{ni2023transformers, grigsby2024amago}, and we have demonstrated that scale-invariant updates can unlock significant improvement in terms of multi-task training. Therefore we argue that the next step is to train this agent in a domain with sufficient task diversity to evaluate meta-learning over $k > 2$ episodes. XLand Minigrid \cite{nikulin2023xland} is a concurrent effort to augment gridworlds with millions of unique tasks and may be a promising domain for future study.

\begin{figure}[h!]
    \vspace{-2mm}
    \centering
    \includegraphics[width=.85\linewidth]{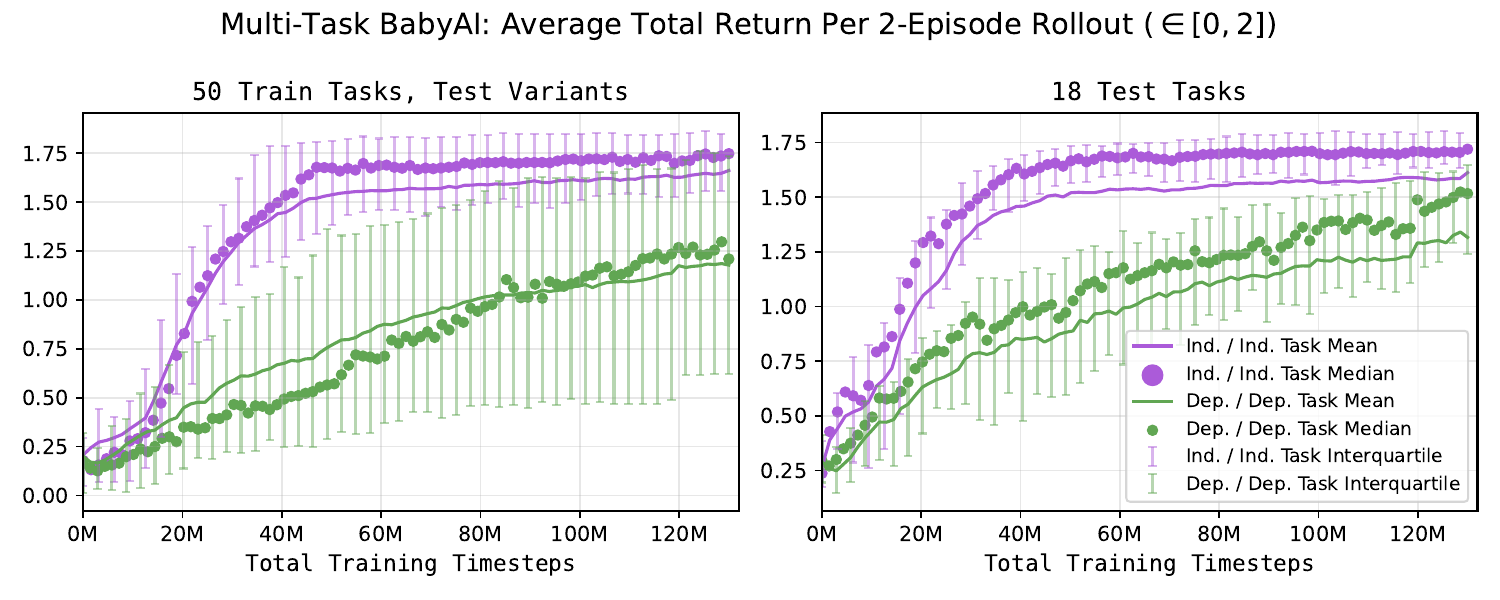}
    \vspace{-4mm}
    \caption{\textbf{Multli-Task BabyAI.} Results are the average over four training seeds and plotted according to the median, mean, and interquartile range over the task set.}
    \label{fig:babyai_main}
    \vspace{-2mm}
\end{figure}
 
\begin{figure}[h!]
    \vspace{-3mm}
    \centering
    \includegraphics[width=.92\linewidth]{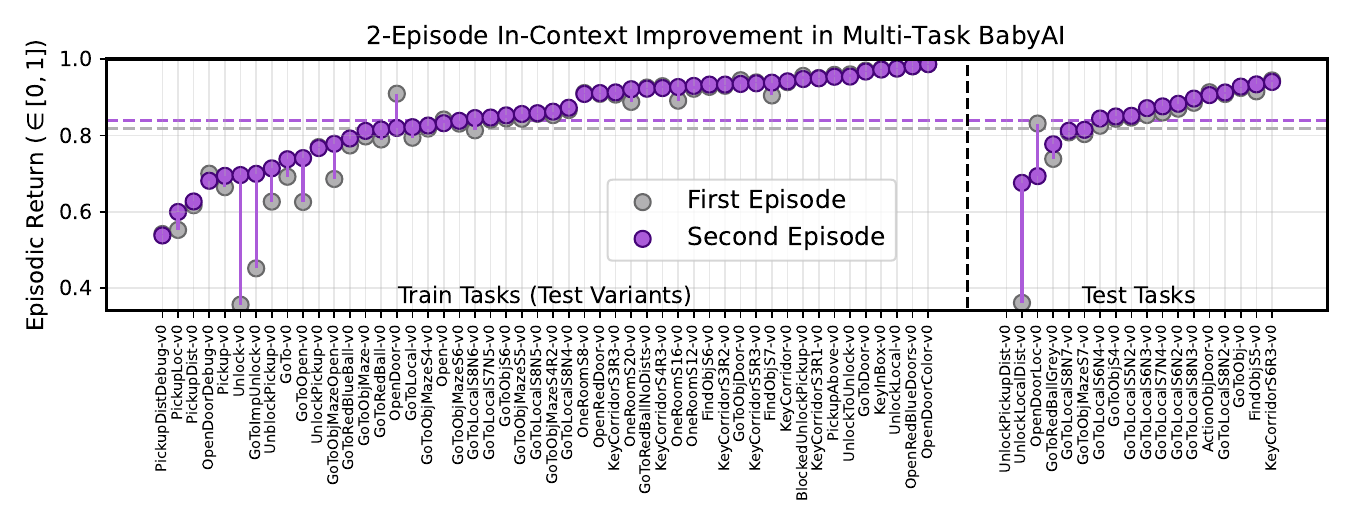}
    \vspace{-4mm}
    \caption{\textbf{In-Context BabyAI.} We measure the average return of ``Ind. / Ind.'' agents by attempt in unseen tasks and/or layouts (variants). Results are arbitrarily sorted in order of increasing second-episode return and are the average of four training seeds and $10$M evaluation timesteps.}
    \label{fig:babyai_icrl}
    \vspace{-6mm}
\end{figure}

\section{Conclusion}
\vspace{-2mm}
Memory-based RL methods are already capable of long-term recall and test-time adaptation to variations of familiar tasks. However, optimization challenges from multi-task learning limit our ability to expand to more general applications. We attribute some, but not all, of this difficulty to the unpredictable way our agents balance the optimization of each task according to their current performance. We show that this bottleneck can be removed without explicit task labels or additional assumptions by simply replacing both the actor and critic objectives with classification terms that do not depend on the current scale of returns. Experiments across meta-learning, generalization, and long-term memory benchmarks find that this update improves multi-task performance without sacrificing sample efficiency. More broadly, these results justify a uniquely accessible approach to adaptive RL. Training a single Transformer on arbitrary recycled data with two loss functions that resemble supervised learning is a simple way to unlock the self-improvement of online RL while sharing intuition and technical details with widely familiar topics like sequence modeling.

\acksection{
This work was partially supported by the National Science Foundation (EFRI-2318065), the Office of Naval Research (N00014-24-1-2550), the DARPA TIAMAT program (HR0011-24-9-0428), JP Morgan, and Sony.}

\bibliography{clean_references}

\begin{thebibliography}{111}
\providecommand{\natexlab}[1]{#1}
\providecommand{\url}[1]{\texttt{#1}}
\expandafter\ifx\csname urlstyle\endcsname\relax
  \providecommand{\doi}[1]{doi: #1}\else
  \providecommand{\doi}{doi: \begingroup \urlstyle{rm}\Url}\fi

\bibitem[Touvron et~al.(2023)Touvron, Martin, Stone, Albert, Almahairi, Babaei, Bashlykov, Batra, Bhargava, Bhosale, et~al.]{touvron2023llama}
Hugo Touvron, Louis Martin, Kevin Stone, Peter Albert, Amjad Almahairi, Yasmine Babaei, Nikolay Bashlykov, Soumya Batra, Prajjwal Bhargava, Shruti Bhosale, et~al.
\newblock Llama 2: Open foundation and fine-tuned chat models.
\newblock \emph{arXiv preprint arXiv:2307.09288}, 2023.

\bibitem[Team et~al.(2023{\natexlab{a}})Team, Anil, Borgeaud, Wu, Alayrac, Yu, Soricut, Schalkwyk, Dai, Hauth, et~al.]{team2023gemini}
Gemini Team, Rohan Anil, Sebastian Borgeaud, Yonghui Wu, Jean-Baptiste Alayrac, Jiahui Yu, Radu Soricut, Johan Schalkwyk, Andrew~M Dai, Anja Hauth, et~al.
\newblock Gemini: a family of highly capable multimodal models.
\newblock \emph{arXiv preprint arXiv:2312.11805}, 2023{\natexlab{a}}.

\bibitem[Hendrycks et~al.(2020)Hendrycks, Burns, Basart, Zou, Mazeika, Song, and Steinhardt]{hendrycks2020measuring}
Dan Hendrycks, Collin Burns, Steven Basart, Andy Zou, Mantas Mazeika, Dawn Song, and Jacob Steinhardt.
\newblock Measuring massive multitask language understanding.
\newblock In \emph{International Conference on Learning Representations}, 2020.

\bibitem[Hendrycks et~al.(2021)Hendrycks, Burns, Kadavath, Arora, Basart, Tang, Song, and Steinhardt]{Hendrycks2021MeasuringMP}
Dan Hendrycks, Collin Burns, Saurav Kadavath, Akul Arora, Steven Basart, Eric Tang, Dawn~Xiaodong Song, and Jacob Steinhardt.
\newblock Measuring mathematical problem solving with the math dataset.
\newblock \emph{ArXiv}, abs/2103.03874, 2021.
\newblock URL \url{https://api.semanticscholar.org/CorpusID:232134851}.

\bibitem[Dua et~al.(2019)Dua, Wang, Dasigi, Stanovsky, Singh, and Gardner]{dua-etal-2019-drop}
Dheeru Dua, Yizhong Wang, Pradeep Dasigi, Gabriel Stanovsky, Sameer Singh, and Matt Gardner.
\newblock {DROP}: A reading comprehension benchmark requiring discrete reasoning over paragraphs.
\newblock In Jill Burstein, Christy Doran, and Thamar Solorio, editors, \emph{Proceedings of the 2019 Conference of the North {A}merican Chapter of the Association for Computational Linguistics: Human Language Technologies, Volume 1 (Long and Short Papers)}, pages 2368--2378, Minneapolis, Minnesota, June 2019. Association for Computational Linguistics.
\newblock \doi{10.18653/v1/N19-1246}.
\newblock URL \url{https://aclanthology.org/N19-1246}.

\bibitem[Brown et~al.(2020)Brown, Mann, Ryder, Subbiah, Kaplan, Dhariwal, Neelakantan, Shyam, Sastry, Askell, et~al.]{brown2020language}
Tom Brown, Benjamin Mann, Nick Ryder, Melanie Subbiah, Jared~D Kaplan, Prafulla Dhariwal, Arvind Neelakantan, Pranav Shyam, Girish Sastry, Amanda Askell, et~al.
\newblock Language models are few-shot learners.
\newblock \emph{Advances in neural information processing systems}, 33:\penalty0 1877--1901, 2020.

\bibitem[Kirsch et~al.(2022)Kirsch, Harrison, Sohl-Dickstein, and Metz]{kirsch2022general}
Louis Kirsch, James Harrison, Jascha Sohl-Dickstein, and Luke Metz.
\newblock General-purpose in-context learning by meta-learning transformers.
\newblock \emph{arXiv preprint arXiv:2212.04458}, 2022.

\bibitem[Beck et~al.(2023)Beck, Vuorio, Liu, Xiong, Zintgraf, Finn, and Whiteson]{beck2023survey}
Jacob Beck, Risto Vuorio, Evan~Zheran Liu, Zheng Xiong, Luisa Zintgraf, Chelsea Finn, and Shimon Whiteson.
\newblock A survey of meta-reinforcement learning.
\newblock \emph{arXiv preprint arXiv:2301.08028}, 2023.

\bibitem[Wang et~al.(2016)Wang, Kurth-Nelson, Tirumala, Soyer, Leibo, Munos, Blundell, Kumaran, and Botvinick]{wang2016learning}
Jane~X Wang, Zeb Kurth-Nelson, Dhruva Tirumala, Hubert Soyer, Joel~Z Leibo, Remi Munos, Charles Blundell, Dharshan Kumaran, and Matt Botvinick.
\newblock Learning to reinforcement learn.
\newblock \emph{arXiv preprint arXiv:1611.05763}, 2016.

\bibitem[Duan et~al.(2016)Duan, Schulman, Chen, Bartlett, Sutskever, and Abbeel]{duan2016rl}
Yan Duan, John Schulman, Xi~Chen, Peter~L Bartlett, Ilya Sutskever, and Pieter Abbeel.
\newblock Rl {$^2$}: Fast reinforcement learning via slow reinforcement learning.
\newblock \emph{arXiv preprint arXiv:1611.02779}, 2016.

\bibitem[Humplik et~al.(2019)Humplik, Galashov, Hasenclever, Ortega, Teh, and Heess]{humplik2019meta}
Jan Humplik, Alexandre Galashov, Leonard Hasenclever, Pedro~A Ortega, Yee~Whye Teh, and Nicolas Heess.
\newblock Meta reinforcement learning as task inference.
\newblock \emph{arXiv preprint arXiv:1905.06424}, 2019.

\bibitem[Ghosh et~al.(2021)Ghosh, Rahme, Kumar, Zhang, Adams, and Levine]{ghosh2021generalization}
Dibya Ghosh, Jad Rahme, Aviral Kumar, Amy Zhang, Ryan~P Adams, and Sergey Levine.
\newblock Why generalization in rl is difficult: Epistemic pomdps and implicit partial observability.
\newblock \emph{Advances in neural information processing systems}, 34:\penalty0 25502--25515, 2021.

\bibitem[Zintgraf(2022)]{zintgraf2022fast}
Luisa Zintgraf.
\newblock \emph{Fast adaptation via meta reinforcement learning}.
\newblock PhD thesis, University of Oxford, 2022.

\bibitem[Zhao et~al.(2022)Zhao, Abbeel, and James]{zhao2022effectiveness}
Mandi Zhao, Pieter Abbeel, and Stephen James.
\newblock On the effectiveness of fine-tuning versus meta-reinforcement learning.
\newblock \emph{Advances in Neural Information Processing Systems}, 35:\penalty0 26519--26531, 2022.

\bibitem[Taiga et~al.(2023)Taiga, Agarwal, Farebrother, Courville, and Bellemare]{taiga2023investigating}
Adrien~Ali Taiga, Rishabh Agarwal, Jesse Farebrother, Aaron Courville, and Marc~G Bellemare.
\newblock Investigating multi-task pretraining and generalization in reinforcement learning.
\newblock In \emph{The Eleventh International Conference on Learning Representations}, 2023.
\newblock URL \url{https://openreview.net/forum?id=sSt9fROSZRO}.

\bibitem[Rusu et~al.(2022)Rusu, Flennerhag, Rao, Pascanu, and Hadsell]{rusu2022probing}
Andrei~Alex Rusu, Sebastian Flennerhag, Dushyant Rao, Razvan Pascanu, and Raia Hadsell.
\newblock Probing transfer in deep reinforcement learning without task engineering.
\newblock In \emph{Conference on Lifelong Learning Agents}, pages 1231--1254. PMLR, 2022.

\bibitem[Yu et~al.(2020{\natexlab{a}})Yu, Quillen, He, Julian, Hausman, Finn, and Levine]{yu2020meta}
Tianhe Yu, Deirdre Quillen, Zhanpeng He, Ryan Julian, Karol Hausman, Chelsea Finn, and Sergey Levine.
\newblock Meta-world: A benchmark and evaluation for multi-task and meta reinforcement learning.
\newblock In \emph{Conference on robot learning}, pages 1094--1100. PMLR, 2020{\natexlab{a}}.

\bibitem[Lee et~al.(2023)Lee, Cho, and Sung]{lee2023parameterizing}
Suyoung Lee, Myungsik Cho, and Youngchul Sung.
\newblock Parameterizing non-parametric meta-reinforcement learning tasks via subtask decomposition.
\newblock In \emph{Thirty-seventh Conference on Neural Information Processing Systems}, 2023.
\newblock URL \url{https://openreview.net/forum?id=JX6UloWrmE}.

\bibitem[Reed et~al.(2022)Reed, Zolna, Parisotto, Colmenarejo, Novikov, Barth-maron, Gim{\'e}nez, Sulsky, Kay, Springenberg, Eccles, Bruce, Razavi, Edwards, Heess, Chen, Hadsell, Vinyals, Bordbar, and de~Freitas]{reed2022a}
Scott Reed, Konrad Zolna, Emilio Parisotto, Sergio~G{\'o}mez Colmenarejo, Alexander Novikov, Gabriel Barth-maron, Mai Gim{\'e}nez, Yury Sulsky, Jackie Kay, Jost~Tobias Springenberg, Tom Eccles, Jake Bruce, Ali Razavi, Ashley Edwards, Nicolas Heess, Yutian Chen, Raia Hadsell, Oriol Vinyals, Mahyar Bordbar, and Nando de~Freitas.
\newblock A generalist agent.
\newblock \emph{Transactions on Machine Learning Research}, 2022.
\newblock ISSN 2835-8856.
\newblock URL \url{https://openreview.net/forum?id=1ikK0kHjvj}.
\newblock Featured Certification, Outstanding Certification.

\bibitem[Hansen et~al.(2023)Hansen, Su, and Wang]{hansen2023td}
Nicklas Hansen, Hao Su, and Xiaolong Wang.
\newblock Td-mpc2: Scalable, robust world models for continuous control.
\newblock In \emph{The Twelfth International Conference on Learning Representations}, 2023.

\bibitem[Gallou{\'e}dec et~al.(2024)Gallou{\'e}dec, Beeching, Romac, and Dellandr{\'e}a]{gallouedec2024jack}
Quentin Gallou{\'e}dec, Edward Beeching, Cl{\'e}ment Romac, and Emmanuel Dellandr{\'e}a.
\newblock Jack of all trades, master of some, a multi-purpose transformer agent.
\newblock \emph{arXiv preprint arXiv:2402.09844}, 2024.

\bibitem[Shaham et~al.(2023)Shaham, Elbayad, Goswami, Levy, and Bhosale]{shaham2023causes}
Uri Shaham, Maha Elbayad, Vedanuj Goswami, Omer Levy, and Shruti Bhosale.
\newblock Causes and cures for interference in multilingual translation.
\newblock In \emph{The 61st Annual Meeting Of The Association For Computational Linguistics}, 2023.

\bibitem[Xin et~al.(2022)Xin, Ghorbani, Gilmer, Garg, and Firat]{xin2022current}
Derrick Xin, Behrooz Ghorbani, Justin Gilmer, Ankush Garg, and Orhan Firat.
\newblock Do current multi-task optimization methods in deep learning even help?
\newblock \emph{Advances in neural information processing systems}, 35:\penalty0 13597--13609, 2022.

\bibitem[Zhang and Yang(2021)]{zhang2021survey}
Yu~Zhang and Qiang Yang.
\newblock A survey on multi-task learning.
\newblock \emph{IEEE Transactions on Knowledge and Data Engineering}, 34\penalty0 (12):\penalty0 5586--5609, 2021.

\bibitem[Hessel et~al.(2019)Hessel, Soyer, Espeholt, Czarnecki, Schmitt, and Van~Hasselt]{hessel2019multi}
Matteo Hessel, Hubert Soyer, Lasse Espeholt, Wojciech Czarnecki, Simon Schmitt, and Hado Van~Hasselt.
\newblock Multi-task deep reinforcement learning with popart.
\newblock In \emph{Proceedings of the AAAI Conference on Artificial Intelligence}, volume~33, pages 3796--3803, 2019.

\bibitem[Vaswani et~al.(2017)Vaswani, Shazeer, Parmar, Uszkoreit, Jones, Gomez, Kaiser, and Polosukhin]{vaswani2017attention}
Ashish Vaswani, Noam Shazeer, Niki Parmar, Jakob Uszkoreit, Llion Jones, Aidan~N Gomez, {\L}ukasz Kaiser, and Illia Polosukhin.
\newblock Attention is all you need.
\newblock \emph{Advances in neural information processing systems}, 30, 2017.

\bibitem[Morad et~al.(2023)Morad, Kortvelesy, Bettini, Liwicki, and Prorok]{morad2023popgym}
Steven Morad, Ryan Kortvelesy, Matteo Bettini, Stephan Liwicki, and Amanda Prorok.
\newblock {POPG}ym: Benchmarking partially observable reinforcement learning.
\newblock In \emph{The Eleventh International Conference on Learning Representations}, 2023.
\newblock URL \url{https://openreview.net/forum?id=chDrutUTs0K}.

\bibitem[Cobbe et~al.(2020)Cobbe, Hesse, Hilton, and Schulman]{cobbe2020leveraging}
Karl Cobbe, Chris Hesse, Jacob Hilton, and John Schulman.
\newblock Leveraging procedural generation to benchmark reinforcement learning.
\newblock In \emph{International conference on machine learning}, pages 2048--2056. PMLR, 2020.

\bibitem[Bellemare et~al.(2013)Bellemare, Naddaf, Veness, and Bowling]{bellemare2013arcade}
Marc~G Bellemare, Yavar Naddaf, Joel Veness, and Michael Bowling.
\newblock The arcade learning environment: An evaluation platform for general agents.
\newblock \emph{Journal of Artificial Intelligence Research}, 47:\penalty0 253--279, 2013.

\bibitem[Chevalier-Boisvert et~al.(2019)Chevalier-Boisvert, Bahdanau, Lahlou, Willems, Saharia, Nguyen, and Bengio]{chevalier-boisvert2018babyai}
Maxime Chevalier-Boisvert, Dzmitry Bahdanau, Salem Lahlou, Lucas Willems, Chitwan Saharia, Thien~Huu Nguyen, and Yoshua Bengio.
\newblock Baby{AI}: First steps towards grounded language learning with a human in the loop.
\newblock In \emph{International Conference on Learning Representations}, 2019.
\newblock URL \url{https://openreview.net/forum?id=rJeXCo0cYX}.

\bibitem[Hafner et~al.(2023)Hafner, Pasukonis, Ba, and Lillicrap]{hafner2023mastering}
Danijar Hafner, Jurgis Pasukonis, Jimmy Ba, and Timothy Lillicrap.
\newblock Mastering diverse domains through world models.
\newblock \emph{arXiv preprint arXiv:2301.04104}, 2023.

\bibitem[Wang et~al.(2020{\natexlab{a}})Wang, Novikov, Zolna, Merel, Springenberg, Reed, Shahriari, Siegel, Gulcehre, Heess, et~al.]{wang2020critic}
Ziyu Wang, Alexander Novikov, Konrad Zolna, Josh~S Merel, Jost~Tobias Springenberg, Scott~E Reed, Bobak Shahriari, Noah Siegel, Caglar Gulcehre, Nicolas Heess, et~al.
\newblock Critic regularized regression.
\newblock \emph{Advances in Neural Information Processing Systems}, 33:\penalty0 7768--7778, 2020{\natexlab{a}}.

\bibitem[Chebotar et~al.(2023)Chebotar, Vuong, Hausman, Xia, Lu, Irpan, Kumar, Yu, Herzog, Pertsch, et~al.]{chebotar2023q}
Yevgen Chebotar, Quan Vuong, Karol Hausman, Fei Xia, Yao Lu, Alex Irpan, Aviral Kumar, Tianhe Yu, Alexander Herzog, Karl Pertsch, et~al.
\newblock Q-transformer: Scalable offline reinforcement learning via autoregressive q-functions.
\newblock In \emph{Conference on Robot Learning}, pages 3909--3928. PMLR, 2023.

\bibitem[Farebrother et~al.(2024)Farebrother, Orbay, Vuong, Ta{\"\i}ga, Chebotar, Xiao, Irpan, Levine, Castro, Faust, et~al.]{farebrother2024stop}
Jesse Farebrother, Jordi Orbay, Quan Vuong, Adrien~Ali Ta{\"\i}ga, Yevgen Chebotar, Ted Xiao, Alex Irpan, Sergey Levine, Pablo~Samuel Castro, Aleksandra Faust, et~al.
\newblock Stop regressing: Training value functions via classification for scalable deep rl.
\newblock \emph{arXiv preprint arXiv:2403.03950}, 2024.

\bibitem[Springenberg et~al.(2024)Springenberg, Abdolmaleki, Zhang, Groth, Bloesch, Lampe, Brakel, Bechtle, Kapturowski, Hafner, et~al.]{springenberg2024offline}
Jost~Tobias Springenberg, Abbas Abdolmaleki, Jingwei Zhang, Oliver Groth, Michael Bloesch, Thomas Lampe, Philemon Brakel, Sarah Bechtle, Steven Kapturowski, Roland Hafner, et~al.
\newblock Offline actor-critic reinforcement learning scales to large models.
\newblock \emph{arXiv preprint arXiv:2402.05546}, 2024.

\bibitem[Chen et~al.(2021{\natexlab{a}})Chen, Lu, Rajeswaran, Lee, Grover, Laskin, Abbeel, Srinivas, and Mordatch]{chen2021decision}
Lili Chen, Kevin Lu, Aravind Rajeswaran, Kimin Lee, Aditya Grover, Misha Laskin, Pieter Abbeel, Aravind Srinivas, and Igor Mordatch.
\newblock Decision transformer: Reinforcement learning via sequence modeling.
\newblock \emph{Advances in neural information processing systems}, 34:\penalty0 15084--15097, 2021{\natexlab{a}}.

\bibitem[Lee et~al.(2022)Lee, Nachum, Yang, Lee, Freeman, Guadarrama, Fischer, Xu, Jang, Michalewski, et~al.]{lee2022multi}
Kuang-Huei Lee, Ofir Nachum, Mengjiao~Sherry Yang, Lisa Lee, Daniel Freeman, Sergio Guadarrama, Ian Fischer, Winnie Xu, Eric Jang, Henryk Michalewski, et~al.
\newblock Multi-game decision transformers.
\newblock \emph{Advances in Neural Information Processing Systems}, 35:\penalty0 27921--27936, 2022.

\bibitem[Hallak et~al.(2015)Hallak, Di~Castro, and Mannor]{hallak2015contextual}
Assaf Hallak, Dotan Di~Castro, and Shie Mannor.
\newblock Contextual markov decision processes.
\newblock \emph{arXiv preprint arXiv:1502.02259}, 2015.

\bibitem[Botvinick et~al.(2019)Botvinick, Ritter, Wang, Kurth-Nelson, Blundell, and Hassabis]{botvinick2019reinforcement}
Matthew Botvinick, Sam Ritter, Jane~X Wang, Zeb Kurth-Nelson, Charles Blundell, and Demis Hassabis.
\newblock Reinforcement learning, fast and slow.
\newblock \emph{Trends in cognitive sciences}, 23\penalty0 (5):\penalty0 408--422, 2019.

\bibitem[Kirk et~al.(2022)Kirk, Zhang, Grefenstette, and Rocktäschel]{kirk2022survey}
Robert Kirk, Amy Zhang, Edward Grefenstette, and Tim Rocktäschel.
\newblock A survey of generalisation in deep reinforcement learning, 2022.

\bibitem[Benjamins et~al.(2022)Benjamins, Eimer, Schubert, Mohan, D{\"o}hler, Biedenkapp, Rosenhahn, Hutter, and Lindauer]{benjamins2022contextualize}
Carolin Benjamins, Theresa Eimer, Frederik Schubert, Aditya Mohan, Sebastian D{\"o}hler, Andr{\'e} Biedenkapp, Bodo Rosenhahn, Frank Hutter, and Marius Lindauer.
\newblock Contextualize me--the case for context in reinforcement learning.
\newblock \emph{arXiv preprint arXiv:2202.04500}, 2022.

\bibitem[Liu et~al.(2021{\natexlab{a}})Liu, Raghunathan, Liang, and Finn]{liu2021decoupling}
Evan~Z Liu, Aditi Raghunathan, Percy Liang, and Chelsea Finn.
\newblock Decoupling exploration and exploitation for meta-reinforcement learning without sacrifices.
\newblock In \emph{International conference on machine learning}, pages 6925--6935. PMLR, 2021{\natexlab{a}}.

\bibitem[Rakelly et~al.(2019)Rakelly, Zhou, Finn, Levine, and Quillen]{rakelly2019efficient}
Kate Rakelly, Aurick Zhou, Chelsea Finn, Sergey Levine, and Deirdre Quillen.
\newblock Efficient off-policy meta-reinforcement learning via probabilistic context variables.
\newblock In \emph{International conference on machine learning}, pages 5331--5340. PMLR, 2019.

\bibitem[Beck et~al.(2024)Beck, Jackson, Vuorio, Xiong, and Whiteson]{beck2024splagger}
Jacob Beck, Matthew Jackson, Risto Vuorio, Zheng Xiong, and Shimon Whiteson.
\newblock Splagger: Split aggregation for meta-reinforcement learning.
\newblock \emph{arXiv preprint arXiv:2403.03020}, 2024.

\bibitem[Laskin et~al.(2022)Laskin, Wang, Oh, Parisotto, Spencer, Steigerwald, Strouse, Hansen, Filos, Brooks, et~al.]{laskin2022context}
Michael Laskin, Luyu Wang, Junhyuk Oh, Emilio Parisotto, Stephen Spencer, Richie Steigerwald, DJ~Strouse, Steven Hansen, Angelos Filos, Ethan Brooks, et~al.
\newblock In-context reinforcement learning with algorithm distillation.
\newblock \emph{arXiv preprint arXiv:2210.14215}, 2022.

\bibitem[Lee et~al.(2024)Lee, Xie, Pacchiano, Chandak, Finn, Nachum, and Brunskill]{lee2024supervised}
Jonathan Lee, Annie Xie, Aldo Pacchiano, Yash Chandak, Chelsea Finn, Ofir Nachum, and Emma Brunskill.
\newblock Supervised pretraining can learn in-context reinforcement learning.
\newblock \emph{Advances in Neural Information Processing Systems}, 36, 2024.

\bibitem[Raparthy et~al.(2023)Raparthy, Hambro, Kirk, Henaff, and Raileanu]{raparthy2023generalization}
Sharath~Chandra Raparthy, Eric Hambro, Robert Kirk, Mikael Henaff, and Roberta Raileanu.
\newblock Generalization to new sequential decision making tasks with in-context learning.
\newblock \emph{arXiv preprint arXiv:2312.03801}, 2023.

\bibitem[Shi et~al.(2024)Shi, Jiang, Grigsby, Fan, and Zhu]{shi2024cross}
Lucy~Xiaoyang Shi, Yunfan Jiang, Jake Grigsby, Linxi Fan, and Yuke Zhu.
\newblock Cross-episodic curriculum for transformer agents.
\newblock \emph{Advances in Neural Information Processing Systems}, 36, 2024.

\bibitem[Mishra et~al.(2017)Mishra, Rohaninejad, Chen, and Abbeel]{mishra2017simple}
Nikhil Mishra, Mostafa Rohaninejad, Xi~Chen, and Pieter Abbeel.
\newblock A simple neural attentive meta-learner.
\newblock \emph{arXiv preprint arXiv:1707.03141}, 2017.

\bibitem[Melo(2022)]{melo2022transformers}
Luckeciano~C Melo.
\newblock Transformers are meta-reinforcement learners.
\newblock In \emph{International Conference on Machine Learning}, pages 15340--15359. PMLR, 2022.

\bibitem[Lu et~al.(2023)Lu, Schroecker, Gu, Parisotto, Foerster, Singh, and Behbahani]{lu2023structured}
Chris Lu, Yannick Schroecker, Albert Gu, Emilio Parisotto, Jakob Foerster, Satinder Singh, and Feryal Behbahani.
\newblock Structured state space models for in-context reinforcement learning.
\newblock \emph{arXiv preprint arXiv:2303.03982}, 2023.

\bibitem[Fakoor et~al.(2019)Fakoor, Chaudhari, Soatto, and Smola]{fakoor2019meta}
Rasool Fakoor, Pratik Chaudhari, Stefano Soatto, and Alexander~J Smola.
\newblock Meta-q-learning.
\newblock \emph{arXiv preprint arXiv:1910.00125}, 2019.

\bibitem[Yang and Nguyen(2021)]{yang2021recurrent}
Zhihan Yang and Hai Nguyen.
\newblock Recurrent off-policy baselines for memory-based continuous control.
\newblock \emph{arXiv preprint arXiv:2110.12628}, 2021.

\bibitem[Ni et~al.(2022)Ni, Eysenbach, and Salakhutdinov]{ni2022recurrent}
Tianwei Ni, Benjamin Eysenbach, and Ruslan Salakhutdinov.
\newblock Recurrent model-free rl can be a strong baseline for many pomdps, 2022.

\bibitem[Team et~al.(2023{\natexlab{b}})Team, Bauer, Baumli, Baveja, Behbahani, Bhoopchand, Bradley-Schmieg, Chang, Clay, Collister, et~al.]{team2023human}
Adaptive~Agent Team, Jakob Bauer, Kate Baumli, Satinder Baveja, Feryal Behbahani, Avishkar Bhoopchand, Nathalie Bradley-Schmieg, Michael Chang, Natalie Clay, Adrian Collister, et~al.
\newblock Human-timescale adaptation in an open-ended task space.
\newblock \emph{arXiv preprint arXiv:2301.07608}, 2023{\natexlab{b}}.

\bibitem[Ni et~al.(2023)Ni, Ma, Eysenbach, and Bacon]{ni2023transformers}
Tianwei Ni, Michel Ma, Benjamin Eysenbach, and Pierre-Luc Bacon.
\newblock When do transformers shine in rl? decoupling memory from credit assignment.
\newblock \emph{arXiv preprint arXiv:2307.03864}, 2023.

\bibitem[Grigsby et~al.(2024)Grigsby, Fan, and Zhu]{grigsby2024amago}
Jake Grigsby, Linxi Fan, and Yuke Zhu.
\newblock {AMAGO}: Scalable in-context reinforcement learning for adaptive agents.
\newblock In \emph{The Twelfth International Conference on Learning Representations}, 2024.
\newblock URL \url{https://openreview.net/forum?id=M6XWoEdmwf}.

\bibitem[Du et~al.(2018)Du, Czarnecki, Jayakumar, Farajtabar, Pascanu, and Lakshminarayanan]{du2018adapting}
Yunshu Du, Wojciech~M Czarnecki, Siddhant~M Jayakumar, Mehrdad Farajtabar, Razvan Pascanu, and Balaji Lakshminarayanan.
\newblock Adapting auxiliary losses using gradient similarity.
\newblock \emph{arXiv preprint arXiv:1812.02224}, 2018.

\bibitem[Yu et~al.(2020{\natexlab{b}})Yu, Kumar, Gupta, Levine, Hausman, and Finn]{yu2020gradient}
Tianhe Yu, Saurabh Kumar, Abhishek Gupta, Sergey Levine, Karol Hausman, and Chelsea Finn.
\newblock Gradient surgery for multi-task learning.
\newblock \emph{Advances in Neural Information Processing Systems}, 33:\penalty0 5824--5836, 2020{\natexlab{b}}.

\bibitem[Li and Gong(2021)]{li2021robust}
Xian Li and Hongyu Gong.
\newblock Robust optimization for multilingual translation with imbalanced data.
\newblock \emph{Advances in Neural Information Processing Systems}, 34:\penalty0 25086--25099, 2021.

\bibitem[Wang et~al.(2020{\natexlab{b}})Wang, Tsvetkov, Firat, and Cao]{wang2020gradient}
Zirui Wang, Yulia Tsvetkov, Orhan Firat, and Yuan Cao.
\newblock Gradient vaccine: Investigating and improving multi-task optimization in massively multilingual models.
\newblock In \emph{International Conference on Learning Representations}, 2020{\natexlab{b}}.

\bibitem[Liu et~al.(2021{\natexlab{b}})Liu, Liu, Jin, Stone, and Liu]{liu2021conflict}
Bo~Liu, Xingchao Liu, Xiaojie Jin, Peter Stone, and Qiang Liu.
\newblock Conflict-averse gradient descent for multi-task learning.
\newblock \emph{Advances in Neural Information Processing Systems}, 34:\penalty0 18878--18890, 2021{\natexlab{b}}.

\bibitem[Royer et~al.(2024)Royer, Blankevoort, and Ehteshami~Bejnordi]{royer2024scalarization}
Amelie Royer, Tijmen Blankevoort, and Babak Ehteshami~Bejnordi.
\newblock Scalarization for multi-task and multi-domain learning at scale.
\newblock \emph{Advances in Neural Information Processing Systems}, 36, 2024.

\bibitem[Kurin et~al.(2022)Kurin, De~Palma, Kostrikov, Whiteson, and Mudigonda]{kurin2022defense}
Vitaly Kurin, Alessandro De~Palma, Ilya Kostrikov, Shimon Whiteson, and Pawan~K Mudigonda.
\newblock In defense of the unitary scalarization for deep multi-task learning.
\newblock \emph{Advances in Neural Information Processing Systems}, 35:\penalty0 12169--12183, 2022.

\bibitem[Sodhani et~al.(2021)Sodhani, Zhang, and Pineau]{sodhani2021multi}
Shagun Sodhani, Amy Zhang, and Joelle Pineau.
\newblock Multi-task reinforcement learning with context-based representations.
\newblock In \emph{International Conference on Machine Learning}, pages 9767--9779. PMLR, 2021.

\bibitem[Kumar et~al.(2022)Kumar, Agarwal, Geng, Tucker, and Levine]{kumar2022offline}
Aviral Kumar, Rishabh Agarwal, Xinyang Geng, George Tucker, and Sergey Levine.
\newblock Offline q-learning on diverse multi-task data both scales and generalizes.
\newblock In \emph{The Eleventh International Conference on Learning Representations}, 2022.

\bibitem[D'Eramo et~al.(2024)D'Eramo, Tateo, Bonarini, Restelli, and Peters]{d2024sharing}
Carlo D'Eramo, Davide Tateo, Andrea Bonarini, Marcello Restelli, and Jan Peters.
\newblock Sharing knowledge in multi-task deep reinforcement learning.
\newblock \emph{arXiv preprint arXiv:2401.09561}, 2024.

\bibitem[garage contributors(2019)]{garage}
The garage contributors.
\newblock Garage: A toolkit for reproducible reinforcement learning research.
\newblock \url{https://github.com/rlworkgroup/garage}, 2019.

\bibitem[Stadie et~al.(2018)Stadie, Yang, Houthooft, Chen, Duan, Wu, Abbeel, and Sutskever]{stadie2018some}
Bradly~C Stadie, Ge~Yang, Rein Houthooft, Xi~Chen, Yan Duan, Yuhuai Wu, Pieter Abbeel, and Ilya Sutskever.
\newblock Some considerations on learning to explore via meta-reinforcement learning.
\newblock \emph{arXiv preprint arXiv:1803.01118}, 2018.

\bibitem[Finn et~al.(2017)Finn, Abbeel, and Levine]{finn2017model}
Chelsea Finn, Pieter Abbeel, and Sergey Levine.
\newblock Model-agnostic meta-learning for fast adaptation of deep networks.
\newblock In \emph{International conference on machine learning}, pages 1126--1135. PMLR, 2017.

\bibitem[Rothfuss et~al.(2018)Rothfuss, Lee, Clavera, Asfour, and Abbeel]{rothfuss2018promp}
Jonas Rothfuss, Dennis Lee, Ignasi Clavera, Tamim Asfour, and Pieter Abbeel.
\newblock Promp: Proximal meta-policy search.
\newblock \emph{arXiv preprint arXiv:1810.06784}, 2018.

\bibitem[Rimon et~al.(2024)Rimon, Jurgenson, Krupnik, Adler, and Tamar]{rimon2024mamba}
Zohar Rimon, Tom Jurgenson, Orr Krupnik, Gilad Adler, and Aviv Tamar.
\newblock {MAMBA}: an effective world model approach for meta-reinforcement learning.
\newblock In \emph{The Twelfth International Conference on Learning Representations}, 2024.
\newblock URL \url{https://openreview.net/forum?id=1RE0H6mU7M}.

\bibitem[Wang et~al.(2021)Wang, King, Porcel, Kurth-Nelson, Zhu, Deck, Choy, Cassin, Reynolds, Song, et~al.]{wang2021alchemy}
Jane~X Wang, Michael King, Nicolas Pierre~Mickael Porcel, Zeb Kurth-Nelson, Tina Zhu, Charlie Deck, Peter Choy, Mary Cassin, Malcolm Reynolds, H~Francis Song, et~al.
\newblock Alchemy: A benchmark and analysis toolkit for meta-reinforcement learning agents.
\newblock In \emph{Thirty-fifth Conference on Neural Information Processing Systems Datasets and Benchmarks Track (Round 2)}, 2021.

\bibitem[Anand et~al.(2022)Anand, Walker, Li, V{\'e}rtes, Schrittwieser, Ozair, Weber, and Hamrick]{anand2022procedural}
Ankesh Anand, Jacob~C Walker, Yazhe Li, Eszter V{\'e}rtes, Julian Schrittwieser, Sherjil Ozair, Theophane Weber, and Jessica~B Hamrick.
\newblock Procedural generalization by planning with self-supervised world models.
\newblock In \emph{International Conference on Learning Representations}, 2022.
\newblock URL \url{https://openreview.net/forum?id=FmBegXJToY}.

\bibitem[Lillicrap et~al.(2015)Lillicrap, Hunt, Pritzel, Heess, Erez, Tassa, Silver, and Wierstra]{lillicrap2015continuous}
Timothy~P Lillicrap, Jonathan~J Hunt, Alexander Pritzel, Nicolas Heess, Tom Erez, Yuval Tassa, David Silver, and Daan Wierstra.
\newblock Continuous control with deep reinforcement learning.
\newblock \emph{arXiv preprint arXiv:1509.02971}, 2015.

\bibitem[Christodoulou(2019)]{christodoulou2019soft}
Petros Christodoulou.
\newblock Soft actor-critic for discrete action settings.
\newblock \emph{arXiv preprint arXiv:1910.07207}, 2019.

\bibitem[Chen et~al.(2021{\natexlab{b}})Chen, Wang, Zhou, and Ross]{chen2021randomized}
Xinyue Chen, Che Wang, Zijian Zhou, and Keith Ross.
\newblock Randomized ensembled double q-learning: Learning fast without a model.
\newblock \emph{arXiv preprint arXiv:2101.05982}, 2021{\natexlab{b}}.

\bibitem[Fedus et~al.(2019)Fedus, Gelada, Bengio, Bellemare, and Larochelle]{fedus2019hyperbolic}
William Fedus, Carles Gelada, Yoshua Bengio, Marc~G Bellemare, and Hugo Larochelle.
\newblock Hyperbolic discounting and learning over multiple horizons.
\newblock \emph{arXiv preprint arXiv:1902.06865}, 2019.

\bibitem[Bellemare et~al.(2023)Bellemare, Dabney, and Rowland]{bellemare2023distributional}
Marc~G Bellemare, Will Dabney, and Mark Rowland.
\newblock \emph{Distributional reinforcement learning}.
\newblock MIT Press, 2023.

\bibitem[Bellemare et~al.(2017)Bellemare, Dabney, and Munos]{bellemare2017distributional}
Marc~G Bellemare, Will Dabney, and R{\'e}mi Munos.
\newblock A distributional perspective on reinforcement learning.
\newblock In \emph{International conference on machine learning}, pages 449--458. PMLR, 2017.

\bibitem[Schrittwieser et~al.(2020)Schrittwieser, Antonoglou, Hubert, Simonyan, Sifre, Schmitt, Guez, Lockhart, Hassabis, Graepel, et~al.]{schrittwieser2020mastering}
Julian Schrittwieser, Ioannis Antonoglou, Thomas Hubert, Karen Simonyan, Laurent Sifre, Simon Schmitt, Arthur Guez, Edward Lockhart, Demis Hassabis, Thore Graepel, et~al.
\newblock Mastering atari, go, chess and shogi by planning with a learned model.
\newblock \emph{Nature}, 588\penalty0 (7839):\penalty0 604--609, 2020.

\bibitem[Hessel et~al.(2021)Hessel, Danihelka, Viola, Guez, Schmitt, Sifre, Weber, Silver, and Van~Hasselt]{hessel2021muesli}
Matteo Hessel, Ivo Danihelka, Fabio Viola, Arthur Guez, Simon Schmitt, Laurent Sifre, Theophane Weber, David Silver, and Hado Van~Hasselt.
\newblock Muesli: Combining improvements in policy optimization.
\newblock In \emph{International conference on machine learning}, pages 4214--4226. PMLR, 2021.

\bibitem[Kapturowski et~al.(2018)Kapturowski, Ostrovski, Quan, Munos, and Dabney]{kapturowski2018recurrent}
Steven Kapturowski, Georg Ostrovski, John Quan, Remi Munos, and Will Dabney.
\newblock Recurrent experience replay in distributed reinforcement learning.
\newblock In \emph{International conference on learning representations}, 2018.

\bibitem[Schulman et~al.(2017)Schulman, Wolski, Dhariwal, Radford, and Klimov]{schulman2017proximal}
John Schulman, Filip Wolski, Prafulla Dhariwal, Alec Radford, and Oleg Klimov.
\newblock Proximal policy optimization algorithms.
\newblock \emph{arXiv preprint arXiv:1707.06347}, 2017.

\bibitem[Abdolmaleki et~al.(2018)Abdolmaleki, Springenberg, Tassa, Munos, Heess, and Riedmiller]{abdolmaleki2018maximum}
Abbas Abdolmaleki, Jost~Tobias Springenberg, Yuval Tassa, Remi Munos, Nicolas Heess, and Martin Riedmiller.
\newblock Maximum a posteriori policy optimisation.
\newblock \emph{arXiv preprint arXiv:1806.06920}, 2018.

\bibitem[Peng et~al.(2019)Peng, Kumar, Zhang, and Levine]{peng2019advantage}
Xue~Bin Peng, Aviral Kumar, Grace Zhang, and Sergey Levine.
\newblock Advantage-weighted regression: Simple and scalable off-policy reinforcement learning.
\newblock \emph{arXiv preprint arXiv:1910.00177}, 2019.

\bibitem[Wang et~al.(2018)Wang, Xiong, Han, Liu, Zhang, et~al.]{wang2018exponentially}
Qing Wang, Jiechao Xiong, Lei Han, Han Liu, Tong Zhang, et~al.
\newblock Exponentially weighted imitation learning for batched historical data.
\newblock \emph{Advances in Neural Information Processing Systems}, 31, 2018.

\bibitem[Chen et~al.(2020)Chen, Zhou, Wang, Wang, Wu, and Ross]{chen2020bail}
Xinyue Chen, Zijian Zhou, Zheng Wang, Che Wang, Yanqiu Wu, and Keith Ross.
\newblock Bail: Best-action imitation learning for batch deep reinforcement learning.
\newblock \emph{Advances in Neural Information Processing Systems}, 33:\penalty0 18353--18363, 2020.

\bibitem[Siegel et~al.(2020)Siegel, Springenberg, Berkenkamp, Abdolmaleki, Neunert, Lampe, Hafner, Heess, and Riedmiller]{siegel2020keep}
Noah~Y Siegel, Jost~Tobias Springenberg, Felix Berkenkamp, Abbas Abdolmaleki, Michael Neunert, Thomas Lampe, Roland Hafner, Nicolas Heess, and Martin Riedmiller.
\newblock Keep doing what worked: Behavioral modelling priors for offline reinforcement learning.
\newblock \emph{arXiv preprint arXiv:2002.08396}, 2020.

\bibitem[Grigsby and Qi(2021)]{grigsby2021closer}
Jake Grigsby and Yanjun Qi.
\newblock A closer look at advantage-filtered behavioral cloning in high-noise datasets.
\newblock \emph{arXiv preprint arXiv:2110.04698}, 2021.

\bibitem[Nair et~al.(2018)Nair, McGrew, Andrychowicz, Zaremba, and Abbeel]{nair2018overcoming}
Ashvin Nair, Bob McGrew, Marcin Andrychowicz, Wojciech Zaremba, and Pieter Abbeel.
\newblock Overcoming exploration in reinforcement learning with demonstrations.
\newblock In \emph{2018 IEEE international conference on robotics and automation (ICRA)}, pages 6292--6299. IEEE, 2018.

\bibitem[Nair et~al.(2020)Nair, Gupta, Dalal, and Levine]{nair2020awac}
Ashvin Nair, Abhishek Gupta, Murtaza Dalal, and Sergey Levine.
\newblock Awac: Accelerating online reinforcement learning with offline datasets.
\newblock \emph{arXiv preprint arXiv:2006.09359}, 2020.

\bibitem[Levine et~al.(2020)Levine, Kumar, Tucker, and Fu]{levine2020offline}
Sergey Levine, Aviral Kumar, George Tucker, and Justin Fu.
\newblock Offline reinforcement learning: Tutorial, review, and perspectives on open problems.
\newblock \emph{arXiv preprint arXiv:2005.01643}, 2020.

\bibitem[Fedus et~al.(2020)Fedus, Ramachandran, Agarwal, Bengio, Larochelle, Rowland, and Dabney]{fedus2020revisiting}
William Fedus, Prajit Ramachandran, Rishabh Agarwal, Yoshua Bengio, Hugo Larochelle, Mark Rowland, and Will Dabney.
\newblock Revisiting fundamentals of experience replay.
\newblock In \emph{International Conference on Machine Learning}, pages 3061--3071. PMLR, 2020.

\bibitem[Janner et~al.(2021)Janner, Li, and Levine]{janner2021reinforcement}
Michael Janner, Qiyang Li, and Sergey Levine.
\newblock Reinforcement learning as one big sequence modeling problem.
\newblock In \emph{ICML 2021 Workshop on Unsupervised Reinforcement Learning}, 2021.

\bibitem[Paster et~al.(2022)Paster, McIlraith, and Ba]{paster2022you}
Keiran Paster, Sheila McIlraith, and Jimmy Ba.
\newblock You can't count on luck: Why decision transformers fail in stochastic environments.
\newblock \emph{arXiv preprint arXiv:2205.15967}, 2022.

\bibitem[Shala et~al.(2024)Shala, Biedenkapp, and Grabocka]{shala2024hierarchical}
Gresa Shala, Andr{\'e} Biedenkapp, and Josif Grabocka.
\newblock Hierarchical transformers are efficient meta-reinforcement learners.
\newblock \emph{arXiv preprint arXiv:2402.06402}, 2024.

\bibitem[Cho et~al.(2014)Cho, van Merrienboer, Gulcehre, Bahdanau, Bougares, Schwenk, and Bengio]{cho2014learning}
Kyunghyun Cho, Bart van Merrienboer, Caglar Gulcehre, Dzmitry Bahdanau, Fethi Bougares, Holger Schwenk, and Yoshua Bengio.
\newblock Learning phrase representations using rnn encoder--decoder for statistical machine translation.
\newblock In \emph{Proceedings of the 2014 Conference on Empirical Methods in Natural Language Processing (EMNLP)}, page 1724. Association for Computational Linguistics, 2014.

\bibitem[Morad et~al.(2024)Morad, Kortvelesy, Liwicki, and Prorok]{morad2024reinforcement}
Steven Morad, Ryan Kortvelesy, Stephan Liwicki, and Amanda Prorok.
\newblock Reinforcement learning with fast and forgetful memory.
\newblock \emph{Advances in Neural Information Processing Systems}, 36, 2024.

\bibitem[Hessel et~al.(2018)Hessel, Modayil, Van~Hasselt, Schaul, Ostrovski, Dabney, Horgan, Piot, Azar, and Silver]{hessel2018rainbow}
Matteo Hessel, Joseph Modayil, Hado Van~Hasselt, Tom Schaul, Georg Ostrovski, Will Dabney, Dan Horgan, Bilal Piot, Mohammad Azar, and David Silver.
\newblock Rainbow: Combining improvements in deep reinforcement learning.
\newblock In \emph{Proceedings of the AAAI conference on artificial intelligence}, volume~32, 2018.

\bibitem[Mnih et~al.(2015)Mnih, Kavukcuoglu, Silver, Rusu, Veness, Bellemare, Graves, Riedmiller, Fidjeland, Ostrovski, et~al.]{mnih2015human}
Volodymyr Mnih, Koray Kavukcuoglu, David Silver, Andrei~A Rusu, Joel Veness, Marc~G Bellemare, Alex Graves, Martin Riedmiller, Andreas~K Fidjeland, Georg Ostrovski, et~al.
\newblock Human-level control through deep reinforcement learning.
\newblock \emph{nature}, 518\penalty0 (7540):\penalty0 529--533, 2015.

\bibitem[Chevalier{-}Boisvert et~al.(2023)Chevalier{-}Boisvert, Dai, Towers, Perez{-}Vicente, Willems, Lahlou, Pal, Castro, and Terry]{MinigridMiniworld23}
Maxime Chevalier{-}Boisvert, Bolun Dai, Mark Towers, Rodrigo Perez{-}Vicente, Lucas Willems, Salem Lahlou, Suman Pal, Pablo~Samuel Castro, and Jordan Terry.
\newblock Minigrid {\&} miniworld: Modular {\&} customizable reinforcement learning environments for goal-oriented tasks.
\newblock In \emph{Advances in Neural Information Processing Systems 36, New Orleans, LA, USA}, December 2023.

\bibitem[Nikulin et~al.(2023)Nikulin, Kurenkov, Zisman, Agarkov, Sinii, and Kolesnikov]{nikulin2023xland}
Alexander Nikulin, Vladislav Kurenkov, Ilya Zisman, Artem Agarkov, Viacheslav Sinii, and Sergey Kolesnikov.
\newblock Xland-minigrid: Scalable meta-reinforcement learning environments in jax.
\newblock \emph{arXiv preprint arXiv:2312.12044}, 2023.

\bibitem[Loshchilov and Hutter(2017)]{loshchilov2017decoupled}
Ilya Loshchilov and Frank Hutter.
\newblock Decoupled weight decay regularization.
\newblock \emph{arXiv preprint arXiv:1711.05101}, 2017.

\bibitem[Shleifer et~al.(2021)Shleifer, Weston, and Ott]{shleifer2021normformer}
Sam Shleifer, Jason Weston, and Myle Ott.
\newblock Normformer: Improved transformer pretraining with extra normalization.
\newblock \emph{arXiv preprint arXiv:2110.09456}, 2021.

\bibitem[Zhai et~al.(2023)Zhai, Likhomanenko, Littwin, Busbridge, Ramapuram, Zhang, Gu, and Susskind]{zhai2023stabilizing}
Shuangfei Zhai, Tatiana Likhomanenko, Etai Littwin, Dan Busbridge, Jason Ramapuram, Yizhe Zhang, Jiatao Gu, and Joshua~M Susskind.
\newblock Stabilizing transformer training by preventing attention entropy collapse.
\newblock In \emph{International Conference on Machine Learning}, pages 40770--40803. PMLR, 2023.

\bibitem[Espeholt et~al.(2018)Espeholt, Soyer, Munos, Simonyan, Mnih, Ward, Doron, Firoiu, Harley, Dunning, et~al.]{espeholt2018impala}
Lasse Espeholt, Hubert Soyer, Remi Munos, Karen Simonyan, Vlad Mnih, Tom Ward, Yotam Doron, Vlad Firoiu, Tim Harley, Iain Dunning, et~al.
\newblock Impala: Scalable distributed deep-rl with importance weighted actor-learner architectures.
\newblock In \emph{International conference on machine learning}, pages 1407--1416. PMLR, 2018.

\bibitem[Wu and He(2018)]{wu2018group}
Yuxin Wu and Kaiming He.
\newblock Group normalization.
\newblock In \emph{Proceedings of the European conference on computer vision (ECCV)}, pages 3--19, 2018.

\bibitem[Yarats et~al.(2021)Yarats, Fergus, Lazaric, and Pinto]{yarats2021mastering}
Denis Yarats, Rob Fergus, Alessandro Lazaric, and Lerrel Pinto.
\newblock Mastering visual continuous control: Improved data-augmented reinforcement learning.
\newblock \emph{arXiv preprint arXiv:2107.09645}, 2021.

\bibitem[Ba et~al.(2016)Ba, Kiros, and Hinton]{ba2016layer}
Jimmy~Lei Ba, Jamie~Ryan Kiros, and Geoffrey~E Hinton.
\newblock Layer normalization.
\newblock \emph{arXiv preprint arXiv:1607.06450}, 2016.

\bibitem[Towers et~al.(2023)Towers, Terry, Kwiatkowski, Balis, Cola, Deleu, Goulão, Kallinteris, KG, Krimmel, Perez-Vicente, Pierré, Schulhoff, Tai, Shen, and Younis]{towers_gymnasium_2023}
Mark Towers, Jordan~K. Terry, Ariel Kwiatkowski, John~U. Balis, Gianluca~de Cola, Tristan Deleu, Manuel Goulão, Andreas Kallinteris, Arjun KG, Markus Krimmel, Rodrigo Perez-Vicente, Andrea Pierré, Sander Schulhoff, Jun~Jet Tai, Andrew Tan~Jin Shen, and Omar~G. Younis.
\newblock Gymnasium, March 2023.
\newblock URL \url{https://zenodo.org/record/8127025}.

\end{thebibliography}
%\printbibliography

\appendix
\newpage

\section{Implementation Details}
\label{app:implementation}

Code for the agent and multi-task environments used in our experiments is available on GitHub at \href{https://github.com/UT-Austin-RPL/amago}{UT-Austin-RPL/amago}.

\paragraph{Base RL Details.} We focus on evaluating changes to the training objective of long-term memory policies trained by off-policy actor-critic updates. Our implementation makes use of many orthogonal technical details from AMAGO \cite{grigsby2024amago}. In the context of this work, the most important detail is the use of a global (task-agnostic) PopArt \cite{hessel2019multi} layer to normalize $Q$-values \textit{of the ``dependent'' update} ($\mathcal{L}_{\text{Actor}}, \mathcal{L}_{\text{Critic}}$), which brings loss functions to a predictable absolute scale but does not change the fact that each task has different relative values. Trying to compare learning updates while skipping this technique would make the $Q$-dependent baselines far too reliant on accurate grid-searches of optimization hyperparameters across domains. Table \ref{tbl:rl_hparams} provides the list of hyperparameters used in our main experiments. All of the results in this paper were completed on NVIDIA A5000 GPUs. We train each agent on one GPU whenever possible but add a second GPU for Procgen Memory-Hard (Figure \ref{fig:procgen_memory}) where model size and context length use all available memory.

\begin{table}[h!]
\resizebox{\textwidth}{!}{\begin{tabular}{@{}lcccccc@{}}
\toprule
                                  & \textbf{Meta-World} & \textbf{POPGym}          & \textbf{Procgen Easy} & \textbf{Procgen Memory-Hard} & \textbf{Atari} & \textbf{BabyAI} \\ \midrule
Learning Rate                     & 5e-4                & 1e-4                     & 1e-4                  & 1e-4                         & 1e-4           & 1e-4                 \\
Batch Size (In Sequences)         & 24                  & 24                       & 24                    & 16                           & 32             & 24                   \\
Replay Buffer Size (in Timesteps) & 8M                  & 15,000 Full Trajectories & 18M                   & 80M                          & 8M            & 10M                  \\
L2 Penalty                        & 1e-4                & 1e-4                     & 5e-3                  & 5e-3                         & 5e-3           & 1e-3                 \\
Grad Clip (Norm)                  & 2                   & 1                        & 1                     & 1                            & 1              & 1                    \\
Learning Update $\gamma$         & \multicolumn{6}{c}{$[0.1, .9, .95, .97, .99, .995, .999]$}                                                                                    \\
Rollout $\gamma$                  & \multicolumn{6}{c}{.999}                                                                                                                      \\
Critic Ensemble Size              & \multicolumn{6}{c}{4}                                                                                                                         \\
Target Network Update $\tau$      & \multicolumn{6}{c}{.003}                                                                                                                      \\ \bottomrule
\end{tabular}}
\caption{\textbf{Learning Hyperparameter Details.}}
\label{tbl:rl_hparams}
\end{table}

\paragraph{Model Architectures.} All of our experiments use causal Transformers trained with the AdamW optimizer \cite{loshchilov2017decoupled}. We apply Normformer \cite{shleifer2021normformer} and $\sigma$Reparam \cite{zhai2023stabilizing} with Leaky ReLU activations to all architectures. These details are intended to stabilize the policy against long training runs of millions of gradient steps but are applicable to any sequence modeling problem; our results did not require any changes that are specific to RL. We use fixed (sinusoidal) position embeddings \cite{vaswani2017attention}. Table \ref{tbl:arch_hparams} lists Transformer architectural details for each of our main experiments. The ``timestep encoder'' maps the potentially multi-modal meta-RL inputs of observations, actions, rewards, and reset signals to a fixed-size vector (Figure \ref{fig:arch}). State-based domains (Meta-World and POPGym) can concatenate and merge this information with a small feed-forward network. In pixel-based environments, we first embed the image observation with a CNN. We evaluated both the small ``Nature'' architecture \cite{mnih2015human} and the residual IMPALA CNN \cite{espeholt2018impala}. Our results default to IMPALA with additional Group Normalization \cite{wu2018group, kumar2022offline}. We apply the random pad and crop data augmentation from DrQV2 \cite{yarats2021mastering} to Procgen and Atari experiments. Image features extracted by the CNN are normalized before being added to action, reward, and terminal data. For consistency with the rest of the architecture we use LayerNorm \cite{ba2016layer}. Preliminary experiments found many of these vision details to be flexible.

\begin{table}[h!]
\resizebox{\textwidth}{!}{\begin{tabular}{@{}lcccccc@{}}
\toprule
                           & \textbf{Meta-World} & \textbf{POPGym} & \textbf{Procgen Easy}                                                              & \textbf{Procgen Memory-Hard}                                                       & \textbf{Atari}                                                                     & \textbf{BabyAI}                                                               \\ \midrule
Timestep Encoder           & FF (512, 256)       & FF (512, 200)   & \begin{tabular}[c]{@{}c@{}}IMPALA CNN\\ {[}16, 32, 32{]} block depths\end{tabular} & \begin{tabular}[c]{@{}c@{}}IMPALA CNN\\ {[}20, 36, 64{]} block depths\end{tabular} & \begin{tabular}[c]{@{}c@{}}IMPALA CNN\\ {[}16, 32, 32{]} block depths\end{tabular} & \begin{tabular}[c]{@{}c@{}}Multi-Modal\\CNN (Grid), RNN (Language), FF (Other) \end{tabular} \\
Transformer Dim.           & 320                 & 256             & 512                                                                                & 512                                                                                & 256                                                                                & 256                                                                                \\
Transformer FF Dim.        & 1280                & 1024            & 2048                                                                               & 2048                                                                               & 1024                                                                               & 1024                                                                               \\
Transformer Layers         & 3                   & 3               & 3                                                                                  & 6                                                                                  & 3                                                                                  & 4                                                                                  \\
Transformer Heads          & 8                   & 8               & 8                                                                                  & 8                                                                                  & 8                                                                                  & 8                                                                                  \\
Context Length (Timesteps) & 256                 & 600             & 128                                                                                & 768                                                                                & 64                                                                                 & 512                                                                                 \\
Actor and Critic MLPs      & (256, 256)          & (256, 256)      & (256, 256)                                                                         & (256, 256)                                                                         & (256, 256)                                                                         & (256, 256)                                                                         \\
Critic Output Bins (Ind.)  & 128                 & 64              & 128                                                                                & 128                                                                                & 128                                                                                & 32                                                                                \\ \bottomrule
\end{tabular}}
\caption{\textbf{Agent Architecture Details.}}
\label{tbl:arch_hparams}
\end{table}

\begin{figure}[h!]
    \centering
    \includegraphics[width=\textwidth]{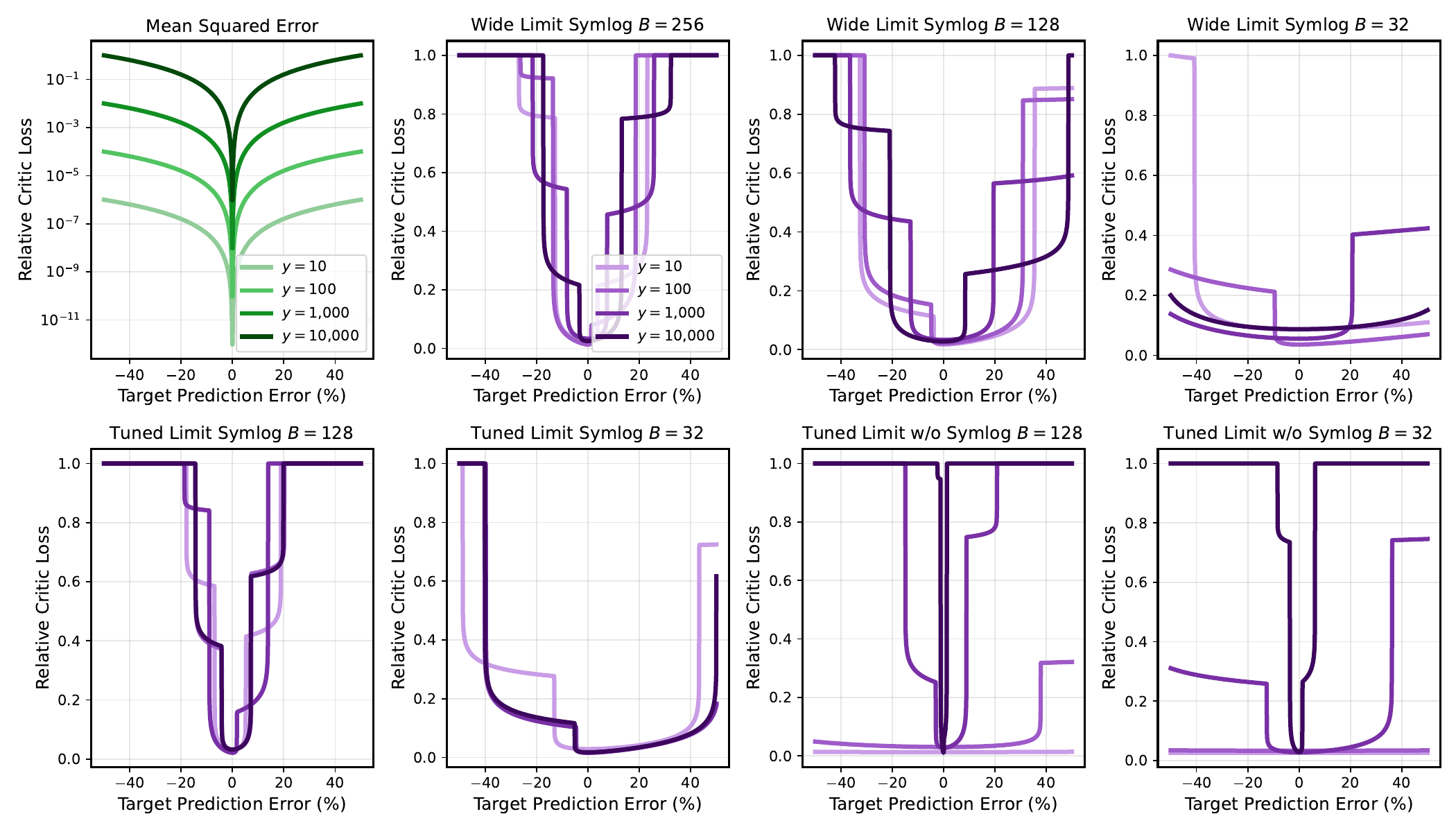}
    \caption{\textbf{Scale-Resistant Regression.} We plot the critic loss as a function of the \textit{relative} prediction error of the value target ($y$) across several orders of magnitude. Standard MSE (top left) assigns significantly different weights to the same relative inaccuracy across tasks according to the absolute value of their current targets. Converting regression to classification can map a wide range of values to a similar loss, but this effect can benefit from tuning. We show $7$ different strategies. ``Wide Limit'' refers to the DreamerV3 approach of setting extreme $(R_\text{low}, R_\text{high})$ defaults that do not need tuning. ``Tuned Limit`` bins set the upper and lower bound close to the expected range for the domain (in this example: $[-100, 15,000]$).}
    \label{fig:relative_losses_extended}
\end{figure}

\paragraph{Value Classification Details.} We convert value regression (Equation \ref{eq:critic_dep}) to classification (Equation \ref{eq:critic_ind}) by creating labels for $B$ return bins $\mathbf{b} = [b_0, b_1, \dots, b_B]$. Bins are typically spaced at fix intervals between pre-defined upper and lower bounds on the return $(b_0 = R_\text{low}, b_B = R_\text{high})$. The critic network ($Q_B$) outputs (softmax) probabilities over these bins, and its value prediction can be recovered by $Q(h_t, a_t) = Q_B(h_t, a_t)^\mathsf{T}\mathbf{b} $.  A scalar temporal difference target $y$ is mapped to a classification label by $\text{twohot}_B: \mathbb{R} \rightarrow [0, 1]^{B}$ --- a function that outputs zeroes everywhere aside from two consecutive indices corresponding to the closest bin below $y$ ($b_{i-1}$) and above $y$ $(b_{i}$), which are filled based on their distance from $y$:

\begin{minipage}{0.45\textwidth}
  \begin{align}
    \text{twohot}_B(y)[i-1] = \frac{y - b_{i-1}}{b_i - b_{i-1}}
    \end{align}
\end{minipage}
\begin{minipage}{0.45\textwidth}
    \begin{align}
    \text{twohot}_B(y)[i] = \frac{b_{i} - y}{b_i - b_{i-1}}
    \end{align}
\end{minipage}

\begin{figure}[h!]
    \centering
    \includegraphics[width=0.8\linewidth]{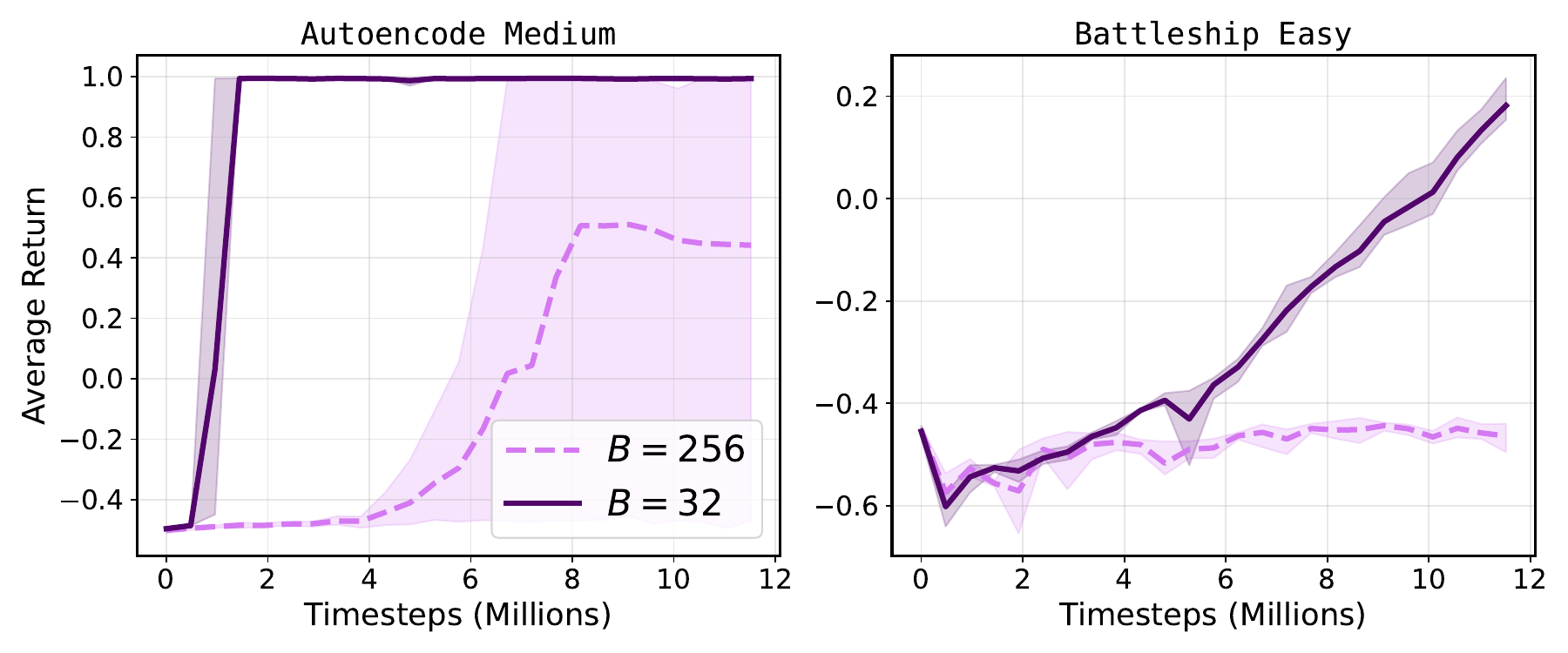}
    \caption{\textbf{Bin Counts in POPGym.} We compare two label space sizes ($B$) with the same $R_\text{low}$ and $R_\text{high}$ limits on two single-task POPGym environments \cite{morad2023popgym}.}
    \label{fig:bin_comparison}
\end{figure}

Therefore there are two main settings to configure for each experiment: 1) the bin limits $[R_\text{low}, R_\text{high}]$, and 2) the number of (evenly spaced) bins between the limits, $B$. We can often set bin limits using prior knowledge of the maximum and minimum return in a given domain, but it can be challenging to establish correct settings for a new application or reward function. The number of bins $B$ defines the ``resolution'' of the critic network. Using fewer bins puts more emphasis on outputting correct label weights while adding more bins shifts focus to outputting correct label indices. This trade-off is interesting because it is somewhat unusual for the label space of a classification problem to be a tunable hyperparameter; for example, the dataset determines the number of labels in object recognition or language modeling. Figure \ref{fig:bin_comparison} provides two single-task examples where lower label counts are more sample efficient, and Figure \ref{fig:procgen_analysis} in the main text supports a similar conclusion in multi-task Procgen. These settings can be easily adjusted in our open-source code, but their trade-offs are underexplored in the results of this work. Instead, we focus on reducing tuning by following details in DreamerV3 \cite{hafner2023mastering}. We set $[R_\text{low}, R_\text{high}]$ extremely wide such that they capture the full range of every experiment and do not need to be tuned. We transform $y$ with ``symlog'' before label creation:
$\text{twohot}_B(y) \leftarrow \text{twohot}_B(\text{symlog(y)})$. Scalar values are recovered from bin probabilities by inverting this transformation with ``symexp'': $Q(h_t, a_t) = \text{symexp}(Q_B(h_t, a_t)^\mathsf{T}\mathbf{b})$, where:

\begin{minipage}{0.45\textwidth}
  \begin{align}
    \text{symlog}(y) = \text{sign}(y) \ln(|y| + 1)
    \end{align}
\end{minipage}
\begin{minipage}{0.45\textwidth}
    \begin{align}
    \text{symexp}(b) = \text{sign}(b) (\exp(|b|) - 1)
    \end{align}
\end{minipage}

\begin{figure}[h!]
    \centering
    \includegraphics[width=0.8\linewidth]{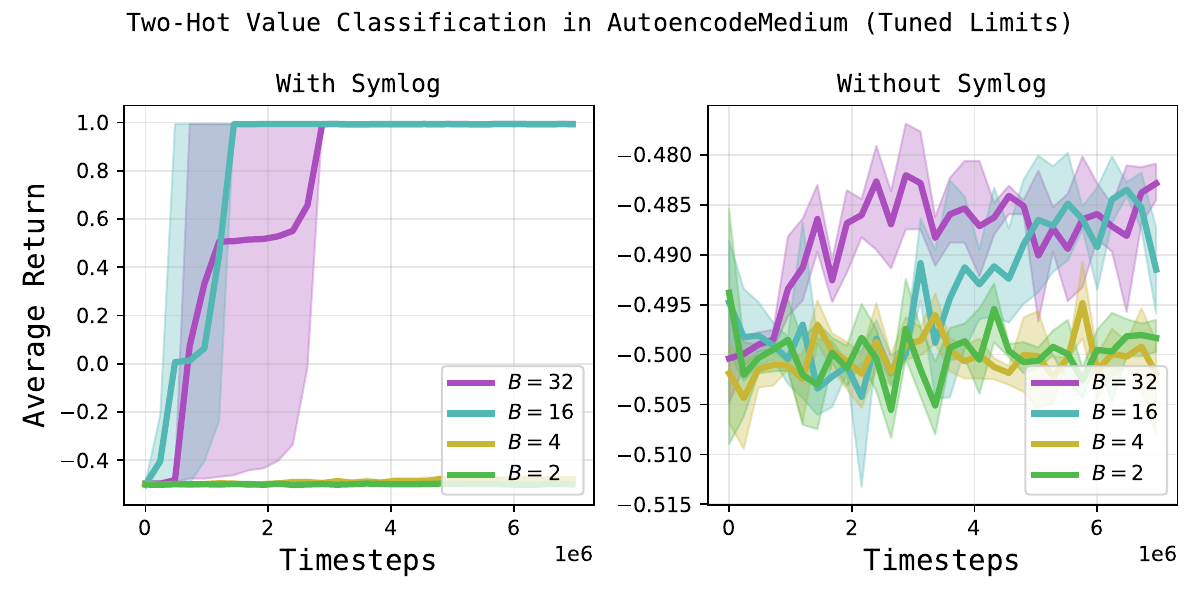}
    \caption{\textbf{Symlog and Bins with Tuned Limits.} We would expect a decrease in the number of value bins to increase sample efficiency up until a point where each bin covers too broad a value range to learn accurate targets. We demonstrate this with a POPGym task \cite{morad2023popgym}. These experiments multiply every reward by $\times 100$ during training and set bin limits of $R_{\text{low}} = -110, R_{\text{high}} = 110$. Shaded areas indicate the maximum and minimum value of three trials.}
    \label{fig:symlog_comparison}
\end{figure}

Bins are spaced evenly between $b_0 = \text{symlog}(R_\text{low})$ and $b_B = \text{symlog}(R_\text{high})$. This symlog trick skews labels to give more resolution to smaller return values while retaining our ability to represent extreme outliers at a lower resolution. In general, we find that setting a wide limit $(R_\text{low},  R_\text{high})$, increasing the label count $B$, and compressing labels with symlog/symexp is the safest way to achieve scale-invariance across domains (Section \ref{sec:method}). An example is illustrated by Figure \ref{fig:relative_losses_extended}. We always use symlog and wide limits of $(R_\text{low} = -1e5, R_\text{high} = 1e5)$, and only tune the bin count $B$ (Table \ref{tbl:arch_hparams}). However, it is likely that sample efficiency can be improved by tuning bin counts, limits, and the use of the symlog transform for each individual domain. Figure \ref{fig:symlog_comparison} illustrates the trade-off between bin counts and symlog with tuned return limits in a single-task setting.

\section{Additional Environment Details}

\paragraph{Multi-Task POPGym.} The environment samples uniformly between $27$ POPGym \cite{morad2023popgym} tasks with discrete actions where both the action space and observation space have dimension $< 30$. A full list is provided as part of Figure \ref{fig:popgym_learning_curves}. The cutoff of dimension $30$ is arbitrary but some limit is necessary because POPGym has several tasks with unusually large action spaces. We are then able to create unified input and output spaces by padding observations and actions to dimensions of $26$. If the agent selects an action index that is not valid for the current task, the environment samples a valid action uniformly at random. An additional observation feature indicates whether the previous action was valid. The one-shot evaluation setting is implemented as in E-RL$^2$ \cite{stadie2018some}, where rewards for the exploration episode are inputs to the policy but are not allowed to become true reward outputs that are optimized by RL updates.

\paragraph{Multi-Game Procgen.} We merge individual Procgen \cite{cobbe2020leveraging} games into a multi-task environment by randomly sampling a new game between meta-rollouts. The same procedurally generated level is then played twice before a new game and level are sampled. In order to allow policies without long-term memory to form accurate value estimates, we draw a small black box in the top left corner of the screen throughout the second episode. We enforce a maximum rollout length (across both episodes) of $H=2500$ for the $5$-game easy mode (Figure \ref{fig:popgym_easy_mode}) and $H=5000$ for the memory-hard mode (Figure \ref{fig:procgen_memory}). In memory-hard mode, we sample levels from the ``memory'' distribution when available and ``hard'' otherwise. The memory distribution extends the hard difficulty by introducing partial observability. Memory mode was not evaluated in the original Procgen results and has been rarely used since. Therefore the results in Figure \ref{fig:procgen_memory} are normalized according to the established scale for hard mode. With the exception of the ``Miner'' game, these upper bound scores transfer to memory mode as it does not adjust the problem difficulty but does hide information about the layout of the level.
\begin{wrapfigure}{r}{.45\textwidth}
\vspace{-12mm}
\begin{center}
    \includegraphics[width=.43\textwidth]{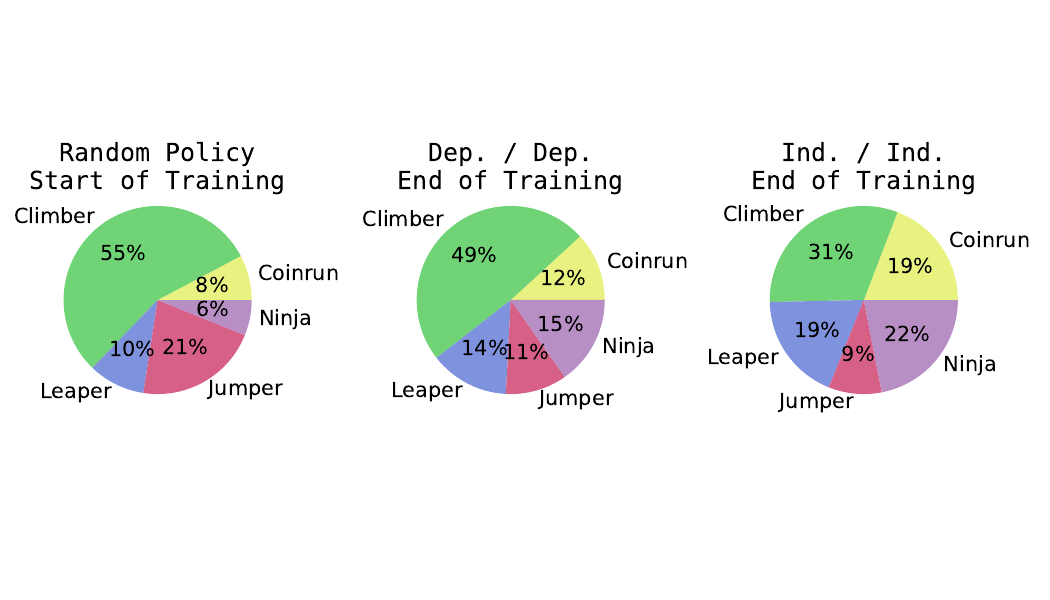}
    \vspace{-12mm}
    \caption{\textbf{Imbalanced Procgen Datasets.} We measure the inflow of experience to the replay buffer by game. Climber accounts for much of our early training data but the buffer is rebalanced as policies improve.}
    \label{fig:procgen_frame_counts}
    \vspace{-8mm}
\end{center}
\end{wrapfigure}

One detail is that selecting a random game between resets does not account for the length of each rollout or the way episode length increases or decreases as the agent improves. As a result, our multi-task Procgen experiments create imbalanced datasets where each game makes up an uneven amount of incoming training data (Figure \ref{fig:procgen_frame_counts}). We consider this to be an interesting additional challenge that off-policy methods with large replay buffers are well suited to address.

\paragraph{Atari.} We select $10$ games that do not involve challenging exploration but that have human-level returns on several different orders of magnitude. The list of games is included in Figure \ref{fig:atari_result}. We use the standardized $\texttt{v5}$ variants of the Atari environments in Gymnasium \cite{towers_gymnasium_2023}. In addition to removing reward clipping, we do not use frame stacking or greyscaling. Instead, agents can learn short-term state estimation and task identification from context sequences of RGB frames.

\paragraph{BabyAI.} BabyAI provides the agent with a $7 \times 7$ partial view of its surroundings in a procedurally generated gridworld. Goals are communicated with relatively simple language instructions, which we represent as sequences from a vocabulary of $29$ tokens. Each reset randomly samples a new goal type and gridworld generation strategy from $50$ tasks in the \href{https://minigrid.farama.org/environments/babyai/}{Minigrid registry} during training, and $18$ tasks during testing. A full list of task names is included as part of Figure \ref{fig:babyai_icrl}.

\section{Additional Results}
\label{app:results}
\label{app:popgym}

\begin{figure}[h!]
    \centering
    \includegraphics[width=\textwidth]{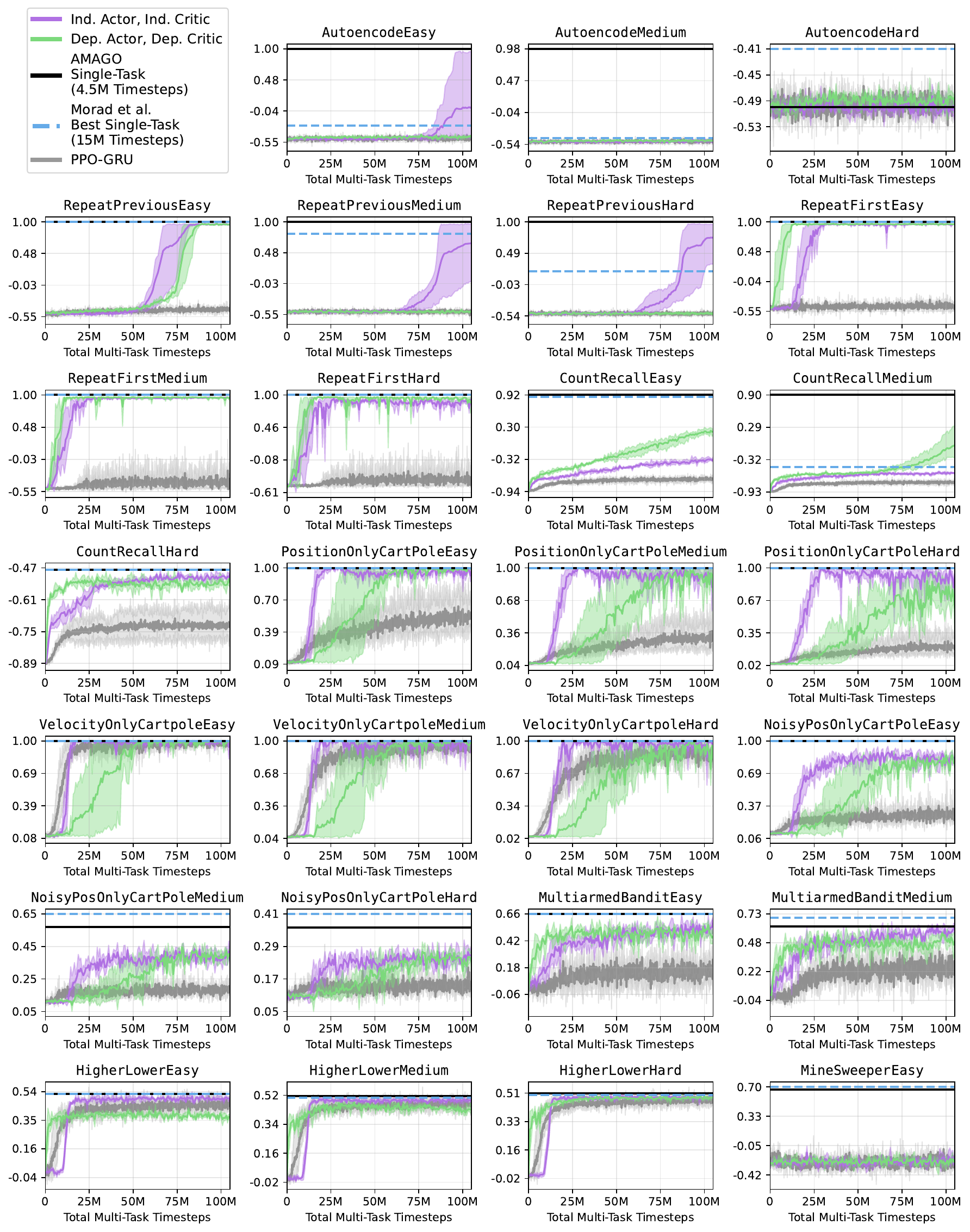}
    \caption{\textbf{One-Shot Multi-Task POPGym.} Plotted on the raw scale of returns in each environment. Error bars denote the maximum and minimum results over three trials. Single-Task reference scores at 15M timesteps indicate the best of 14 sequence model backbones trained by PPO in the results of Morad et al. \cite{morad2024reinforcement}. AMAGO \cite{grigsby2024amago} single-task reference scores are taken at $4.5$M timesteps, which is a safe upper-bound on the timesteps encountered for each task during multi-task training.}
    \label{fig:popgym_learning_curves}
\end{figure}

\begin{figure}[h!]
    \centering
    \includegraphics[width=\linewidth]{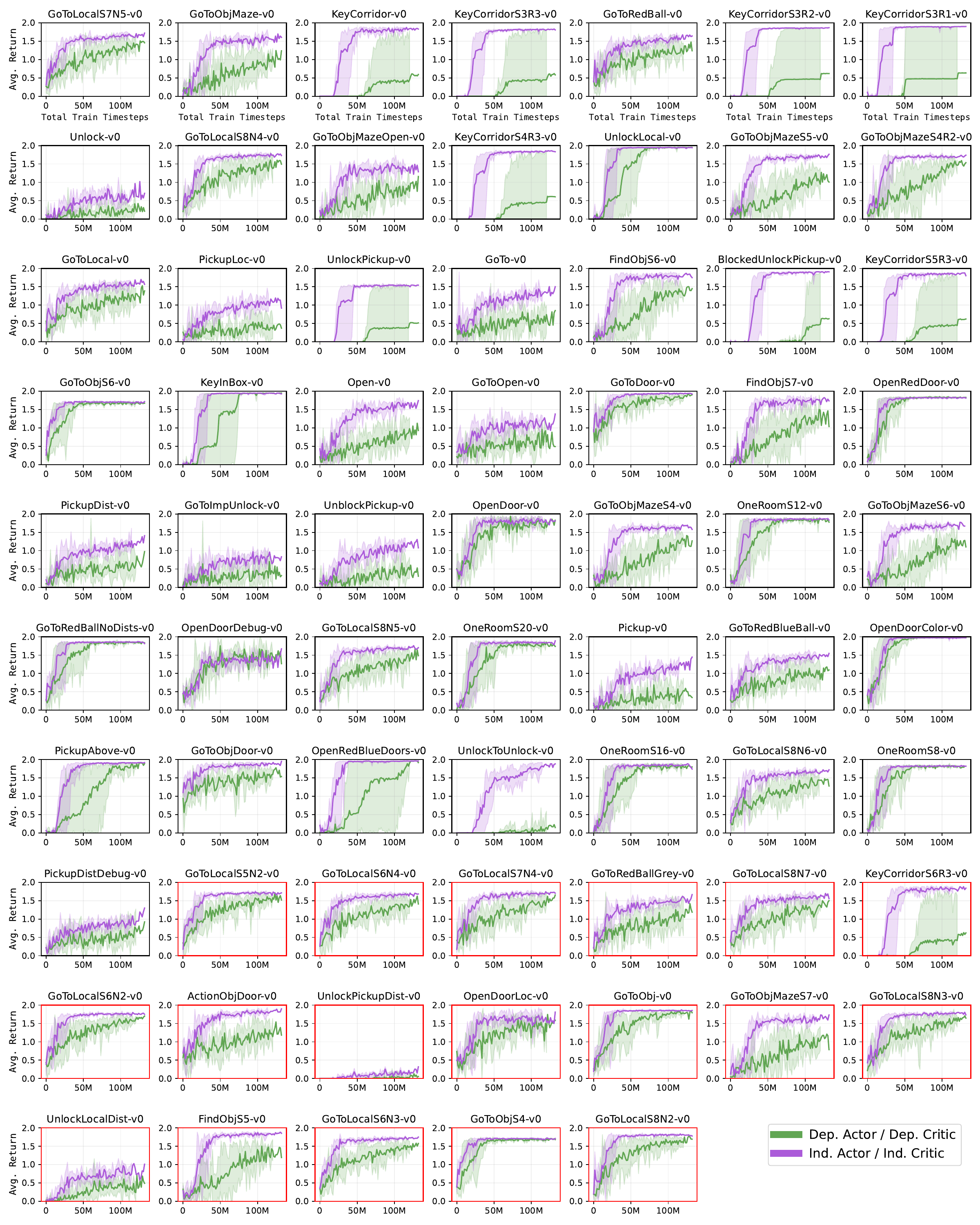}
    \caption{\textbf{Two-Episode Multi-Task BabyAI.} Held-out test tasks are highlighted in \textcolor{red}{red}. Error bars denote the maximum and minimum value of four random trials.}
    \label{fig:babyai_appendix}
\end{figure}

Large figures displaying results from each task in our POPGym (Figure \ref{fig:popgym_learning_curves}) and BabyAI (Figure \ref{fig:babyai_appendix}) experiments are listed on the following two pages. These figures are followed by learning curves for each Meta-World ML45 task.

\begin{figure}[h!]
    \centering
    \includegraphics[width=300px]{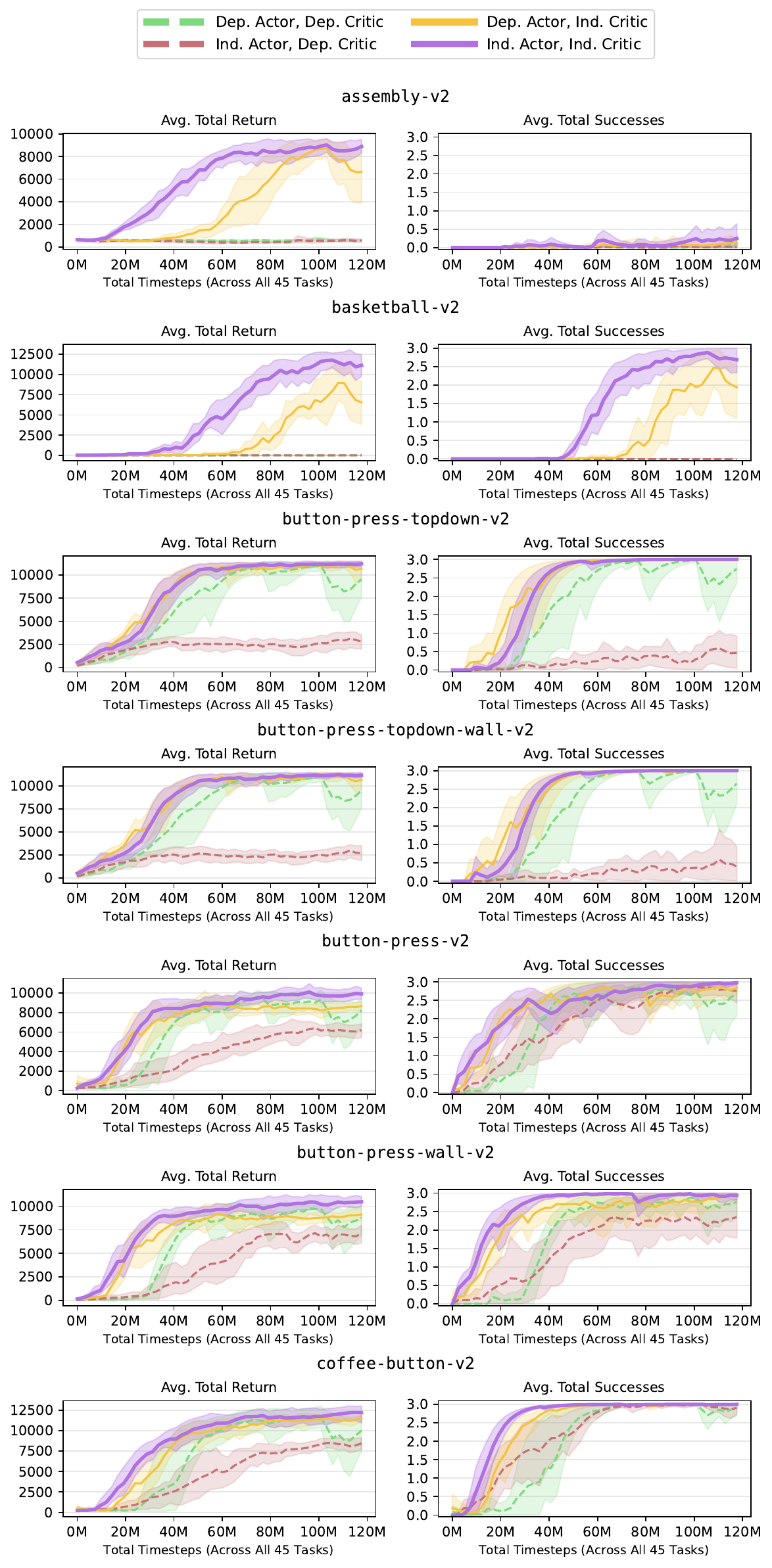}
    \caption{\textbf{Three-Episode Meta-World ML45 Learning Curves (Part 1/7)}}
\end{figure}
\newpage

\begin{figure}[h!]
    \centering
    \includegraphics[width=300px]{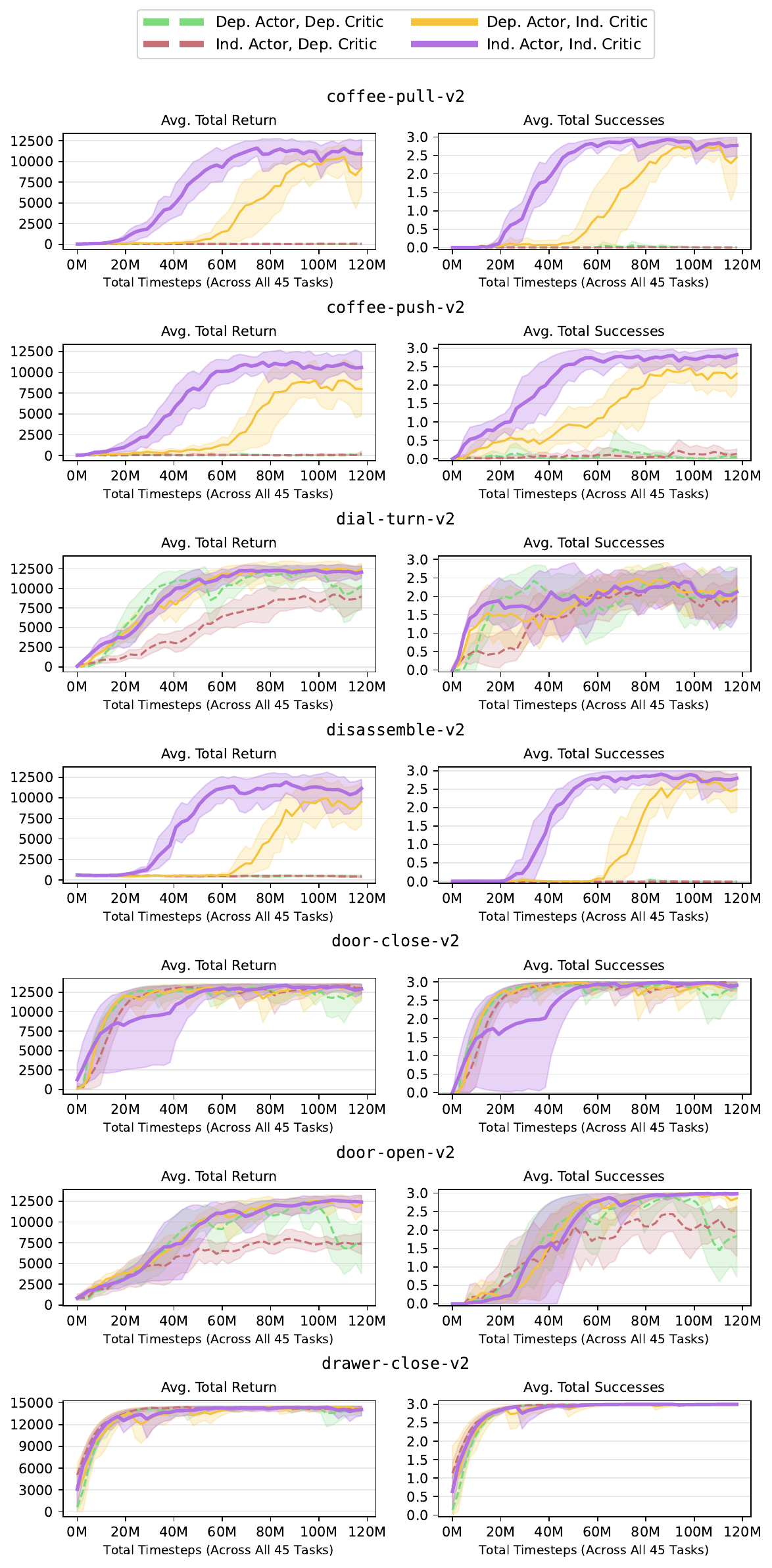}
    \caption{\textbf{Three-Episode Meta-World ML45 Learning Curves (Part 2/7)}}
\end{figure}
\newpage

\begin{figure}[h!]
    \centering
    \includegraphics[width=300px]{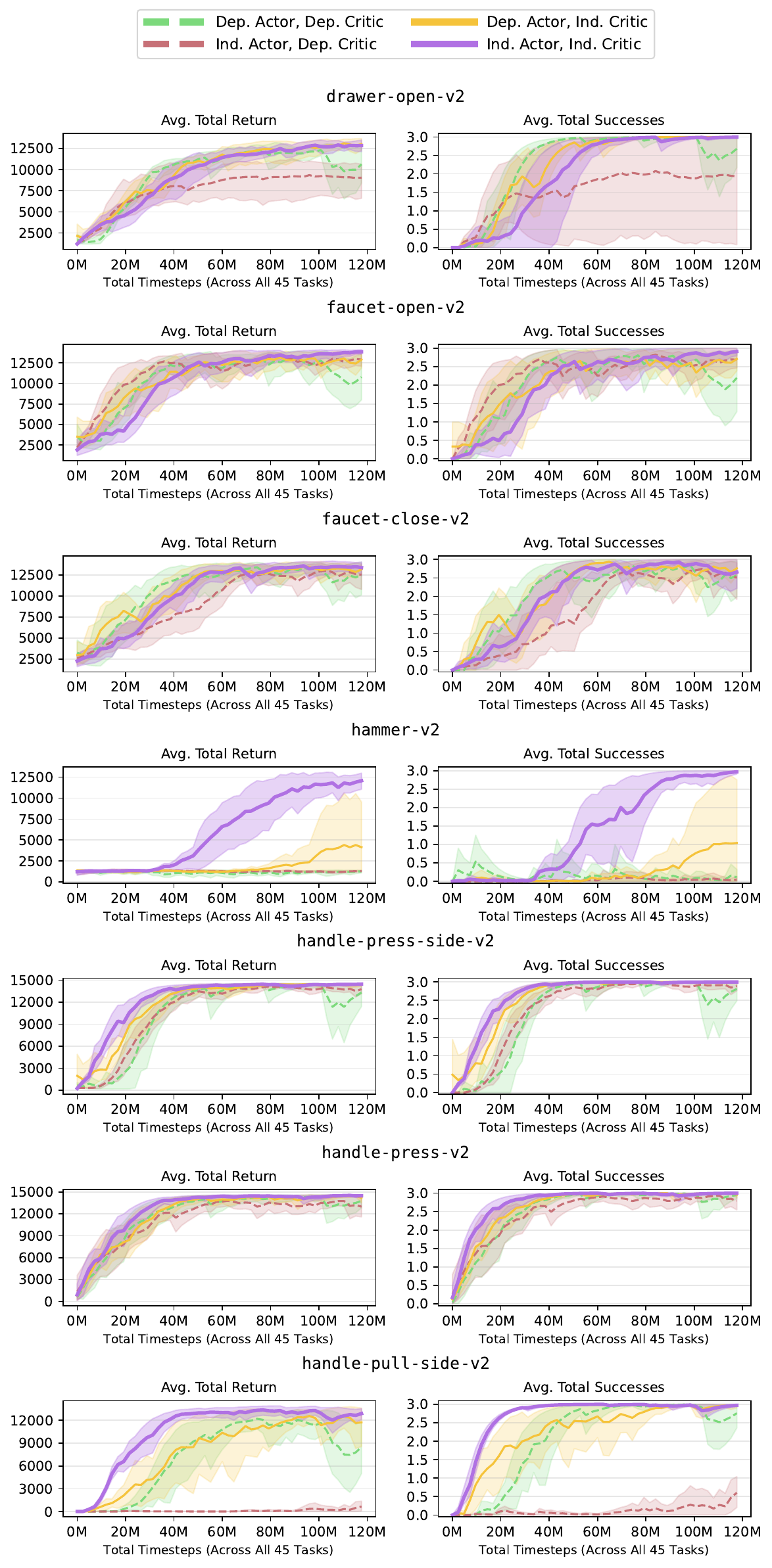}
    \caption{\textbf{Three-Episode Meta-World ML45 Learning Curves (Part 3/7)}}
\end{figure}
\newpage

\begin{figure}[h!]
    \centering
    \includegraphics[width=300px]{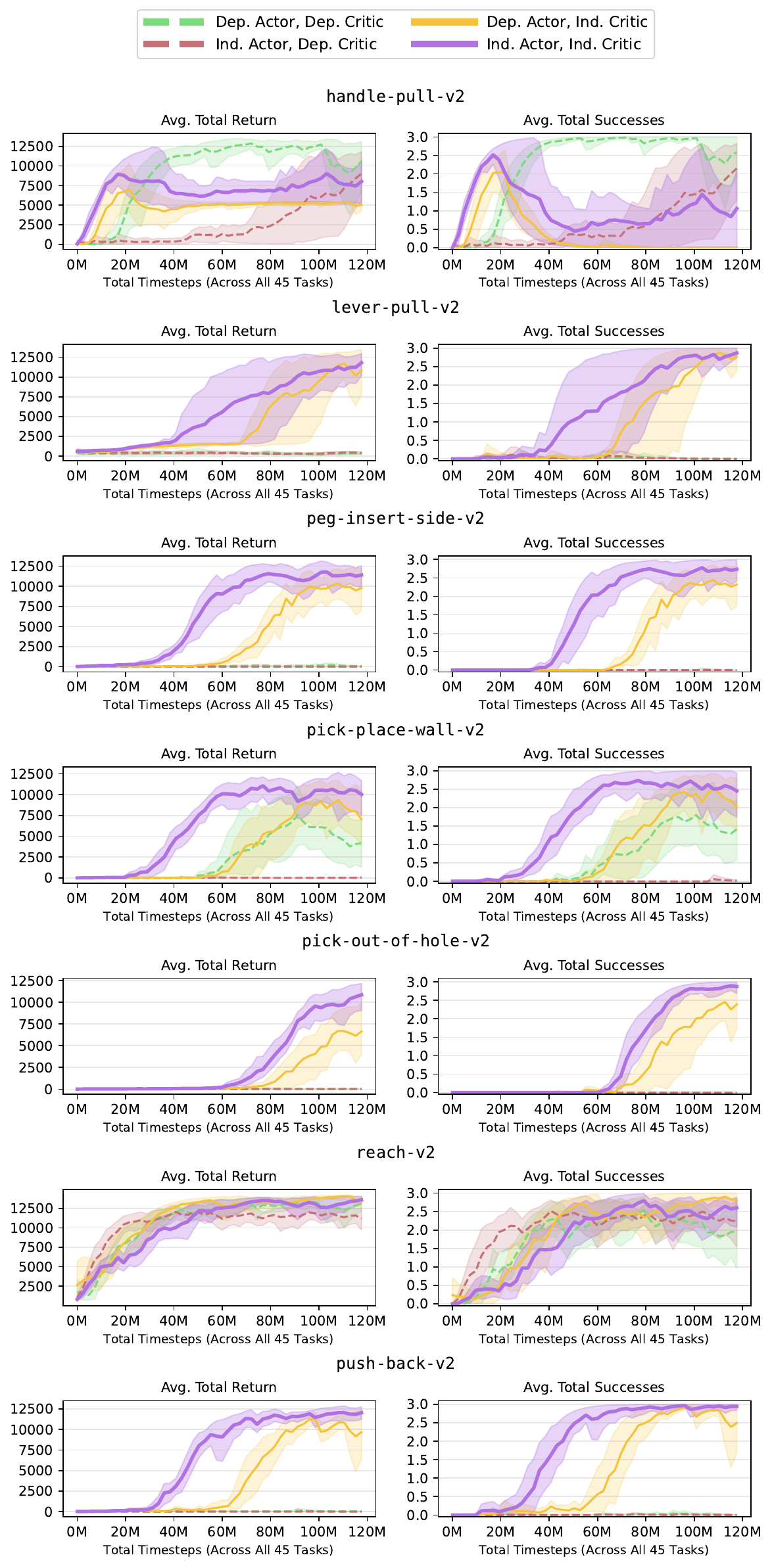}
    \caption{\textbf{Three-Episode Meta-World ML45 Learning Curves (Part 4/7)}}
\end{figure}
\newpage

\begin{figure}[h!]
    \centering
    \includegraphics[width=300px]{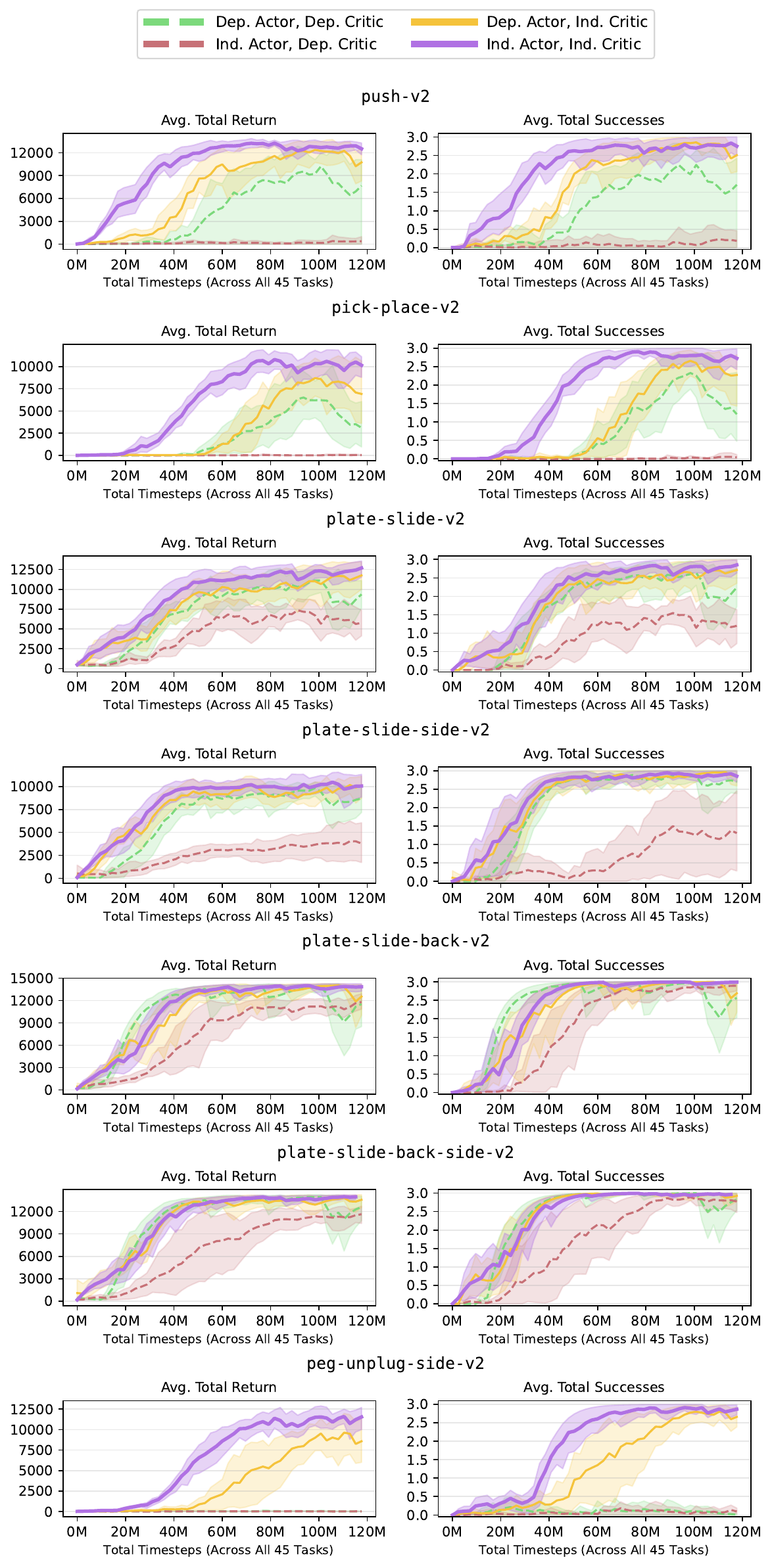}
    \caption{\textbf{Three-Episode Meta-World ML45 Learning Curves (Part 5/7)}}
\end{figure}
\newpage

\begin{figure}[h!]
    \centering
    \includegraphics[width=300px]{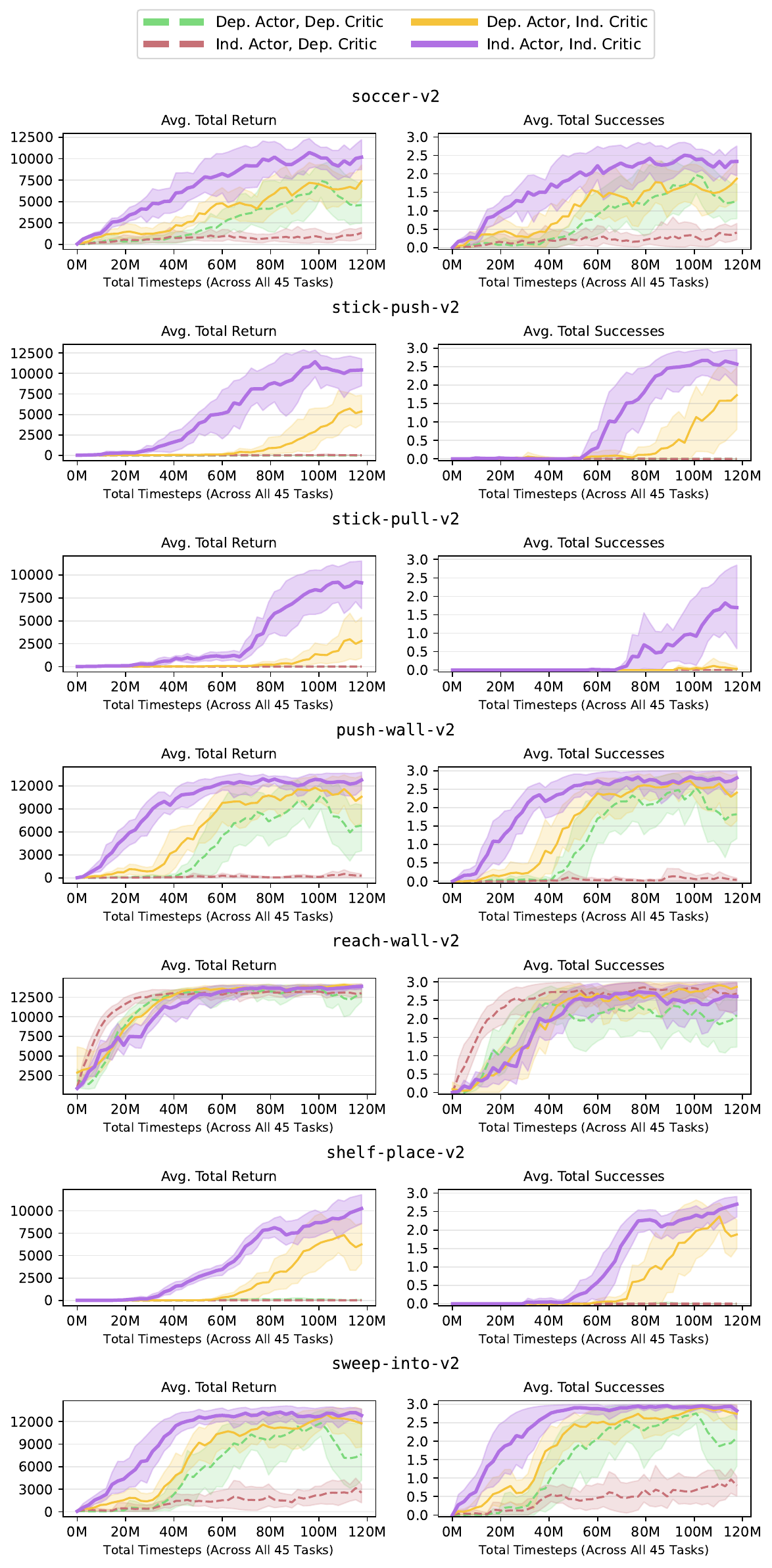}
    \caption{\textbf{Three-Episode Meta-World ML45 Learning Curves (Part 6/7)}}
\end{figure}
\newpage

\begin{figure}[h!]
    \centering
    \includegraphics[width=330px]{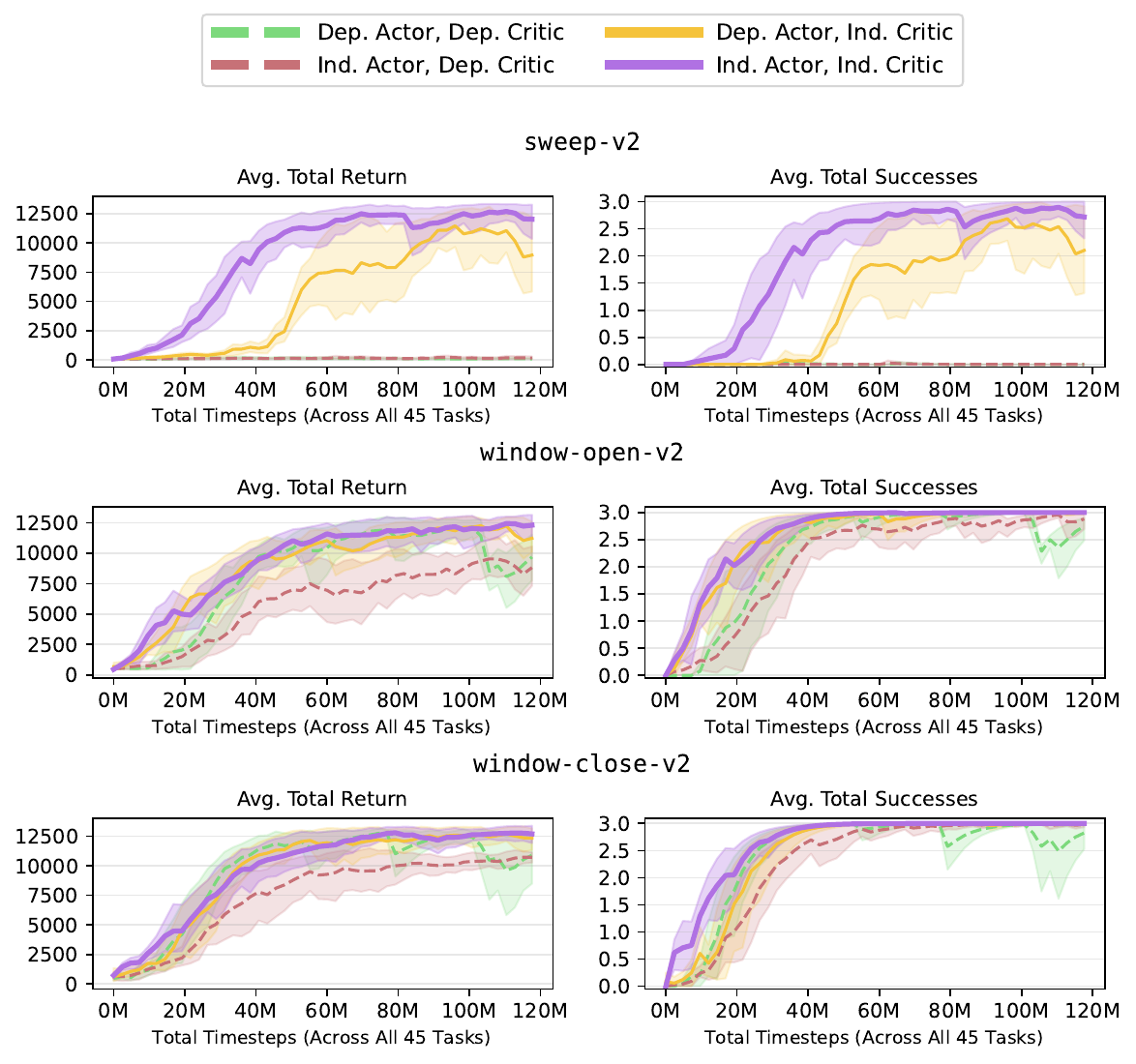}
    \caption{\textbf{Three-Episode Meta-World ML45 Learning Curves (Part 7/7)}}
\end{figure}
\newpage

\clearpage
\end{document}